\documentclass{article}

\usepackage{htlab}
\usepackage[numbers]{natbib}

\labname{HiveTraceLab}
\labnamepos{left}

\usepackage[utf8]{inputenc}
\usepackage[T1]{fontenc}
\usepackage{hyperref}
\usepackage{hyperxmp}
\hypersetup{colorlinks,allcolors=black}
\usepackage{xurl}
\usepackage{booktabs}
\usepackage{amsfonts}
\usepackage{nicefrac}
\usepackage{microtype}
\usepackage{xcolor}

\usepackage{graphicx}
\usepackage{subcaption}
\usepackage{float}
\usepackage{wrapfig}
\usepackage{placeins}

\usepackage{amssymb}

\usepackage{amsmath,amsfonts,bm}

\def\eqref#1{equation~\ref{#1}}

\def\1{\bm{1}}

\DeclareMathAlphabet{\mathsfit}{\encodingdefault}{\sfdefault}{m}{sl}
\SetMathAlphabet{\mathsfit}{bold}{\encodingdefault}{\sfdefault}{bx}{n}

\usepackage{array}
\usepackage{booktabs}
\usepackage{multirow}
\usepackage{tabularx}
\usepackage{longtable}
\usepackage{makecell}
\usepackage{siunitx}

\usepackage{url}
\usepackage{tabularx}
\usepackage{booktabs}
\usepackage{wrapfig}
\usepackage{graphicx}
\usepackage{multirow}

\usepackage{array}
\usepackage{caption}
\usepackage{siunitx}

\title{Cross-Lingual Jailbreak Detection via Semantic Codebooks}

\author{
Shirin Alanova \\
AI Talent Hub, ITMO University \\
\email{shirin.alanova@gmail.com}
\And
Bogdan Minko \\
HiveTraceLab \\
\email{minkobogdan2001@gmail.com}
\And
Sabrina Sadiekh \\
HiveTraceLab \\
\email{sadsobr7@gmail.com}
\And
Evgeniy Kokuykin \\
HiveTraceLab \\
\email{evgeniy.kokuykin@raftds.com}
}

\labname{HiveTraceLab}
\lablogo{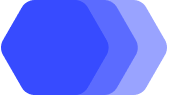}
\lablogowidth{1.3cm}
\labnamepos{left}

\begin{document}
\maketitle

\begin{abstract}
Safety mechanisms for large language models (LLMs) remain predominantly English-centric, creating systematic vulnerabilities in multilingual deployment. Prior work shows that translating malicious prompts into other languages can substantially increase jailbreak success rates, exposing a structural cross-lingual security gap. We investigate whether such attacks can be mitigated through language-agnostic semantic similarity without retraining or language-specific adaptation. Our approach compares multilingual query embeddings against a fixed English codebook of jailbreak prompts, operating as a training-free external guardrail for black-box LLMs.
We conduct a systematic evaluation across four languages, two translation pipelines, four safety benchmarks, three embedding models, and three target LLMs (Qwen, Llama, GPT-3.5). Our results reveal two distinct regimes of cross-lingual transfer. On curated benchmarks containing canonical jailbreak templates, semantic similarity generalizes reliably across languages, achieving near-perfect separability (AUC up to 0.99) and substantial reductions in absolute attack success rates under strict low–false-positive constraints. 
However, under distribution shift—on behaviorally diverse and heterogeneous unsafe benchmarks—separability degrades markedly (AUC $\approx$  0.60-0.70), and recall in the security-critical low-FPR regime drops across all embedding models.
\end{abstract}
\section{Introduction}
\label{sec:introduction}

Safety alignment for large language models (LLMs) remains predominantly English-centric. A recent large-scale analysis of nearly 300 publications (2020–2024) by Yong et al.~\cite{yong2025state} reveals a substantial linguistic imbalance: non-English languages are underrepresented by a factor of 5–10× even among high-resource languages such as Chinese and Arabic, while low-resource languages are largely ignored. Crucially, non-English settings are rarely studied as standalone security scenarios, but instead treated as peripheral extensions of English evaluations. This imbalance creates structural blind spots in multilingual deployment, where safety mechanisms may fail outside the linguistic domain in which they were primarily aligned.
Empirical evidence confirms this vulnerability. Li et al.\cite{li2024crosslanguageinvestigationjailbreakattacks} demonstrate that translating malicious prompts into other languages systematically increases jailbreak success rates across advanced LLMs. Similarly, Shen et al.\cite{shen-etal-2024-language} show that refusal behaviors learned in English often fail to transfer reliably across languages. While fine-tuning on multilingual jailbreak datasets can substantially reduce attack success rates, such approaches require model access, additional data curation, and language-specific adaptation, limiting their applicability in black-box or resource-constrained settings.
These findings raise a practical question: can cross-lingual jailbreak attempts be mitigated without retraining, additional data collection, or language-specific tuning?

In this work, we investigate whether harmful intent can be detected through language-agnostic semantic similarity after translation. Specifically, we ask:

\textit{Can a fixed English semantic codebook detect cross-lingual jailbreak attempts through multilingual embeddings, without retraining or translation-based preprocessing?}

To answer this question, we construct a training-free detection framework that compares multilingual query embeddings against a fixed English codebook of jailbreak prompts. We systematically evaluate this approach across four languages, two translation pipelines, four safety benchmarks, three embedding models, and three target LLMs.

Our contributions are as follows:

\begin{itemize}
    \item We propose a training-free cross-lingual guardrail based on semantic similarity to a fixed English codebook, deployable as an external filter for black-box LLMs
    \item We provide a systematic empirical evaluation of cross-lingual transfer across multiple benchmarks, identifying settings where similarity-based detection generalizes reliably and settings where performance degrades under distribution shift
    \item We analyze performance in the security-critical low-FPR regime and demonstrate that embedding choice critically affects detection robustness
    \item We position semantic codebooks as a lightweight first-line filter for multilingual safety, highlighting both their practical utility and their limitations
\end{itemize}

\section{Related Works}

\subsection{\textbf{Cross-lingual security gaps}}  

LLM security remains extremely English-centric.  Yong et al~\cite{yong2025state}, confirmed that over 90\% of publications ignore non-English languages even in multilingual models. This has real-world implications: Li et al.~\cite{li2024crosslanguageinvestigationjailbreakattacks} demonstrate that translating jailbreak requests into other languages systematically bypasses filters even in GPT-4 and LLaMA. Shen et al.~\cite{shen-etal-2024-language} confirm that models trained to reject harmful requests in English often fail in other languages.

\subsection{\textbf{Attack detection methods}}

Existing approaches are divided into internal and external. Internal methods (RLHF, DPO, PEFT) require access to the model and retraining~\cite{ouyang2022rlhf, rafailov2024dpo, hu2021lora}, which is not applicable to black-box APIs. External methods work as preprocessing, but most of them are based on English: rule-based filters can be readily bypassed by rephrasing~\cite{li2026matterssafetyalignment}, and classifiers such as Prompt-Guard-86M~\cite{meta_prompt_guard_86m} lose their effectiveness on translated queries~\cite{shen-etal-2024-language}.

\subsection{\textbf{Semantic similarity as a signal}}

A more promising approach is to compare query embeddings with a codebook of known attacks using cosine similarity. Hypothesis: malicious intent forms a connected region in semantic space, which allows detecting zero-shot attacks when using aligned multilingual embeddings~\cite{michail2025examiningmultilingualembeddingmodels}. This is empirically confirmed: ~\cite{galinkin2024improved} and ~\cite{ayub2024embeddingbasedclassifiersdetectprompt} show that frozen embeddings with classical ML classifiers outperform fine-tuned transformers at FPR $<$ 1\%.

\subsection{\textbf{The role of embedders}}

Not all multilingual models are the same. Bell et al.~\cite{bell2025} find that the “translate → classify” strategy outperforms direct multilingual classifiers in toxicity tasks, especially for low-resource languages. However, for jailbreaks, semantic drift during translation (e.g., via Google Translate) remains a problem. BGE-M3~\cite{bge-m3} demonstrates state-of-the-art alignment, but its effectiveness for security in realistic settings has not been tested. 

Theoretically, cross-language translation should be supported if the embedding spaces are well-aligned, that is, if they exhibit isomorphism, isometry, and isotropy~\cite{pallucchini2025lost}. However, in practice, even the most powerful models suffer from translation artifacts, especially in the case of typologically distant languages such as Arabic or Chinese. Alternative architectures, such as SPIRE~\cite{segal2025spire}, attempt to address this issue by dynamically indexing conflicting fragments, but they remain sensitive to tokenization and require manual curation.

Our work fills this critical gap. We systematically evaluate whether a fixed English semantic code book combined with BGE-M3 can reliably prevent cross-language jailbreaks -- without retraining, without language-specific tuning, and without dependence on translation quality. This is not just another detector, it is an attempt to address the fundamental blindness of LLMs to their own vulnerabilities outside of English.
\section{Methodology}

\subsection{Detection Framework Overview}

\begin{figure*}[ht]
\centering
\includegraphics[width=0.9\textwidth]{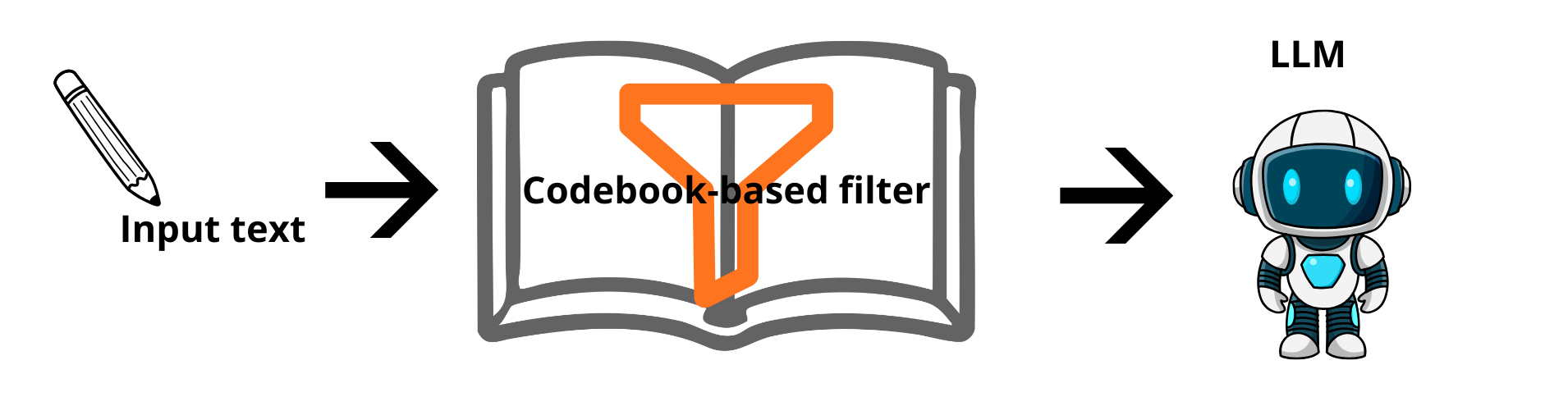}
\caption{
Overview of the proposed cross-lingual semantic filtering framework. 
Incoming user input (in any language) is encoded using a multilingual embedding model and compared against a fixed English codebook of jailbreak prompts. 
If the maximum cosine similarity exceeds a predefined threshold, the query is blocked; otherwise, it is forwarded to the target LLM. 
The approach operates as a training-free external guardrail and does not require translation or model fine-tuning.
}
\label{fig:main_pipeline}
\end{figure*}

Our approach is a training-free cross-lingual guardrail based on semantic similarity to a fixed English codebook of jailbreak prompts. Given an incoming query in any language, we encode it using a multilingual sentence embedding model and compare it to a set of pre-computed embeddings of unsafe English prompts. Detection is performed via nearest-neighbor similarity without retraining, translation-based preprocessing, or language-specific adaptation. Figure~\ref{fig:main_pipeline} illustrates the overall detection pipeline.

\subsection{Codebook Construction}

The codebook is a fixed set of English-language unsafe prompts embedded into a shared vector space.

We construct the codebook from the training split of \textit{jayavibhav/prompt-injection-safety}. Starting from 15,237 unsafe examples, we remove duplicates, extremely short prompts, and malformed inputs. To reduce annotation noise, each candidate prompt is additionally checked by two external guard models (Prompt-Guard-86M~\cite{meta_prompt_guard_86m} and Qwen3Guard-Gen-4B~\cite{zhao2025qwen3guard}). A prompt is retained if at least one model confirms it as unsafe.

The final codebook contains 13,811 unique jailbreak prompts. The codebook remains fixed throughout all experiments.

\subsection{Embedding and Detection Rule}

All prompts are encoded using multilingual sentence embeddings (BGE-M3 unless otherwise stated). Let $f(x) \in \mathbb{R}^{1024}$ denote the L2-normalized embedding of query $x$.

Let $\mathcal{C} = \{\mathbf{c}_1, \dots, \mathbf{c}_N\}$ be the codebook embeddings. For a query $x$, we compute the maximum cosine similarity:

\[
s(x) = \max_{\mathbf{c} \in \mathcal{C}} f(x)^\top \mathbf{c}.
\]

The query is classified as unsafe if $s(x) \ge \tau$, where $\tau$ is selected on validation data.

This yields a nearest-neighbor similarity detector operating entirely in embedding space and requiring no model fine-tuning.

The threshold $\tau$ is selected on a held-out validation partition derived from the training split of jayavibhav/prompt-injection-safety. Specifically, we perform a grid search over candidate thresholds to maximize TPR subject to the hard operational constraint FPR$\leq 1\%$. This optimization protocol reflects deployment settings where false positives incur disproportionate filtering costs. The resulting per-language thresholds and their corresponding operating points are reported in Appendix~\ref{app:tau_tables}.

\subsection{Evaluation Metrics}

We report standard binary classification metrics (TPR, FPR, AUC). Because deployment-grade safety systems must operate under strict false-positive constraints, we focus particularly on the low-FPR regime (e.g., TPR at FPR $\leq 1\%$).

To assess real-world mitigation impact, we additionally measure Attack Success Rate (ASR), defined as the proportion of unsafe prompts that elicit harmful outputs from a target LLM. We report both baseline ASR (without filtering) and ASR after applying the semantic filter.

\subsection{Cross-Lingual Setup}

To simulate realistic cross-lingual attacks, English prompts are translated into Russian, Chinese, and Arabic using two translation pipelines: (i) Google Translate (commercial system), (ii) M2M100 (facebook/m2m100\_418M).

No post-editing or back-translation is applied. This reflects realistic attacker behavior and preserves potential translation artifacts.

\subsection{Evaluation Datasets}

We evaluate our approach on four publicly available binary-labeled safety datasets covering prompt injection, jailbreak behaviors, and broader unsafe content:

\begin{itemize}
    \item \textbf{jayavibhav/prompt-injection-safety\cite{jayavibhav2024promptinjection}} (used for codebook construction and testing),
    \item \textbf{xTRam1/safe-guard-prompt-injection\cite{li2025injecguardbenchmark}},
    \item \textbf{JailbreakBench/JBB-Behaviors\cite{chao2024jailbreakbench}},
    \item \textbf{nvidia/Aegis-AI-Content-Safety-Dataset-2.0\cite{ghosh2024data}}.
\end{itemize}

For large datasets, we sample balanced subsets of 1,000 prompts (500 safe / 500 unsafe). JailbreakBench (200 prompts) is used in full as a stress-test benchmark. Evaluation sets are not manually re-cleaned beyond deduplication in order to preserve realistic noise conditions.

\subsection{Benchmark Characteristics}

The four benchmarks differ in structure and difficulty.

Benchmarks 1–2 correspond to \texttt{jayavibhav/prompt-injection-safety} and \texttt{xTRam1/safe-guard-prompt-injection}, respectively, and primarily contain prompt-injection and canonical jailbreak-style instructions with relatively homogeneous attack patterns. These settings resemble templated or pattern-driven attacks.

In contrast, \texttt{JailbreakBench} (Benchmark 3) and \texttt{Aegis-AI-Content-Safety-Dataset-2.0} (Benchmark 4) exhibit substantially greater behavioral diversity. They include heterogeneous unsafe categories (e.g., harmful instructions, sensitive topics, adversarial rephrasings) and contain prompts that do not necessarily follow canonical jailbreak templates. This increased diversity and distributional variability makes similarity-based detection more challenging and provides a stress-test for cross-lingual transfer.

We report results separately across benchmarks to capture these differences in structure and difficulty.
\section{Experiments and Results}
\label{sec:experiments}

We structure the evaluation around three research questions. Across all experiments, a consistent pattern emerges: similarity-based detection exhibits two distinct behavioral regimes depending on benchmark structure. On curated benchmarks dominated by canonical jailbreak templates, cross-lingual transfer is strong and stable. In contrast, on behaviorally diverse and heterogeneous unsafe benchmarks, performance degrades substantially. The following research questions quantify this effect at three levels: separability, operating constraints, and end-to-end mitigation.

\textbf{RQ1 (Cross-lingual separability).}
Does an English-only semantic codebook transfer across languages in embedding space, i.e., can it separate unsafe vs.\ safe prompts after translation?

\textbf{RQ2 (Security-critical operating point).}
Is there a practical single-threshold operating regime with low false-positive rate (FPR) that still catches a meaningful fraction of attacks across languages and benchmarks?

\textbf{RQ3 (End-to-end mitigation).}
Does the detector reduce jailbreak success when used as an external pre-filter for black-box LLMs, and how does this depend on the target model and translation method?

\paragraph{RQ1: Cross-lingual separability.}

\begin{figure}[ht]
\centering
\begin{subfigure}{0.31\textwidth}
  \includegraphics[width=\linewidth]{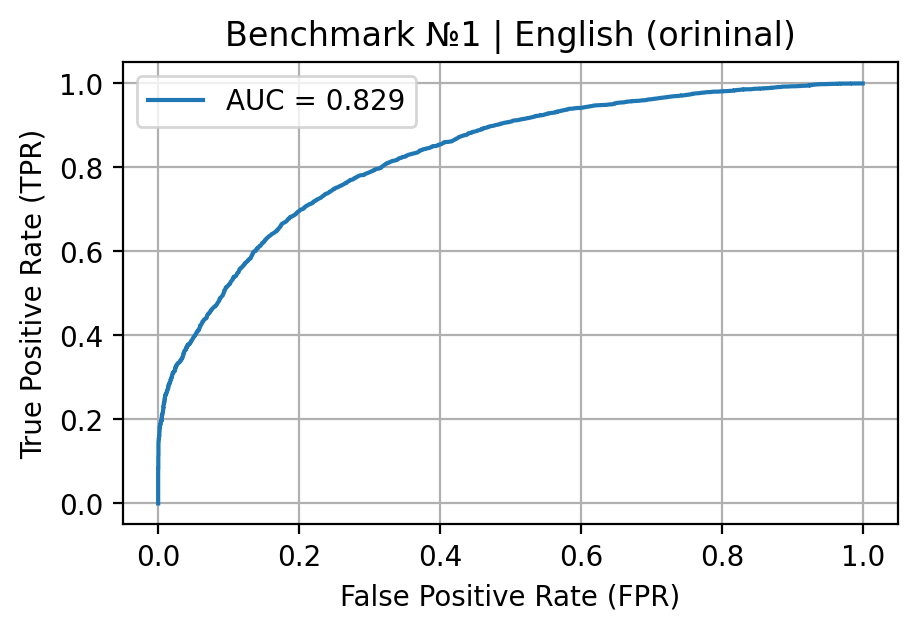}
\end{subfigure}\hfill
\begin{subfigure}{0.31\textwidth}
  \includegraphics[width=\linewidth]{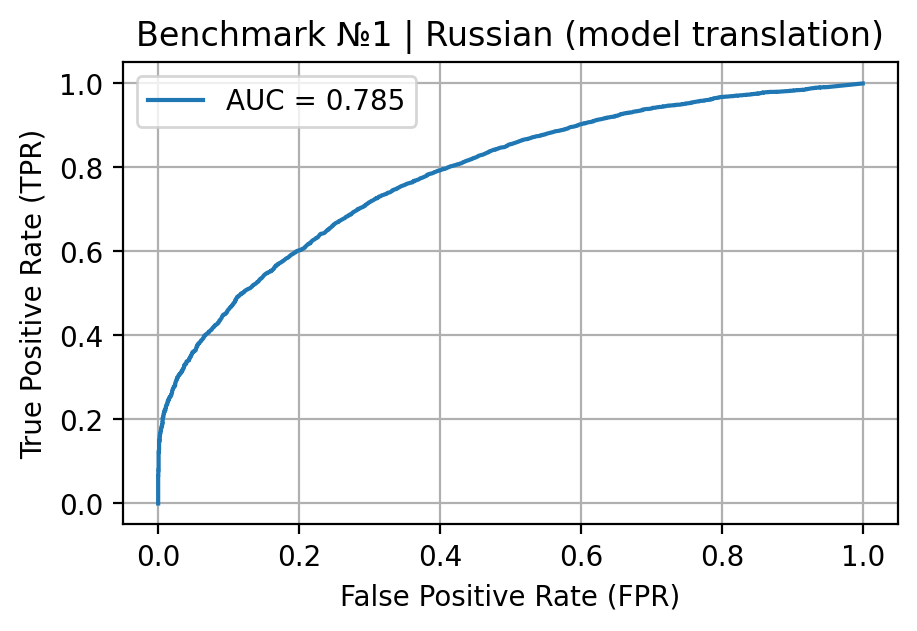}
\end{subfigure}\hfill
\begin{subfigure}{0.31\textwidth}
  \includegraphics[width=\linewidth]{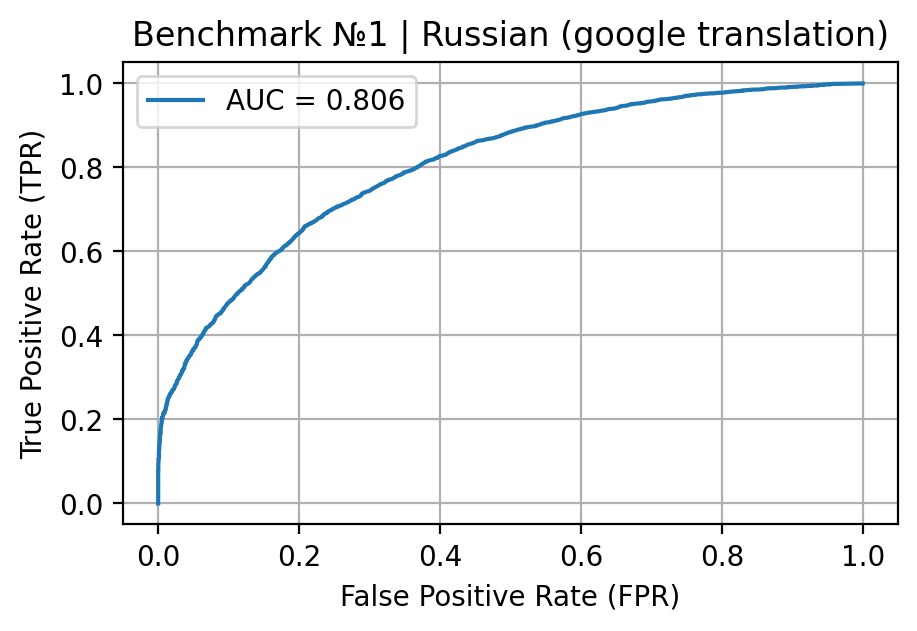}
\end{subfigure}

\medskip

\begin{subfigure}{0.31\textwidth}
  \includegraphics[width=\linewidth]{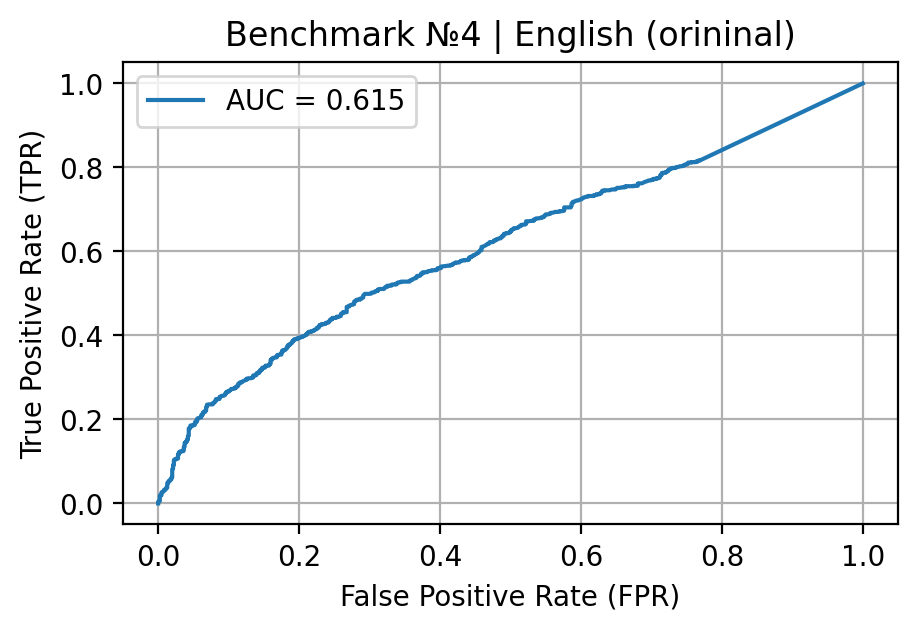}
\end{subfigure}\hfill
\begin{subfigure}{0.31\textwidth}
  \includegraphics[width=\linewidth]{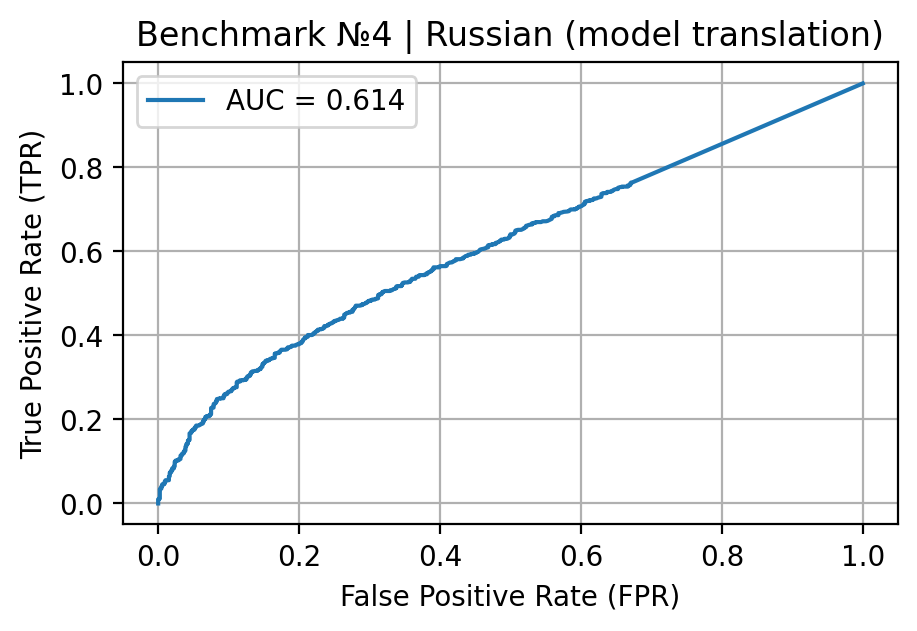}
\end{subfigure}\hfill
\begin{subfigure}{0.31\textwidth}
  \includegraphics[width=\linewidth]{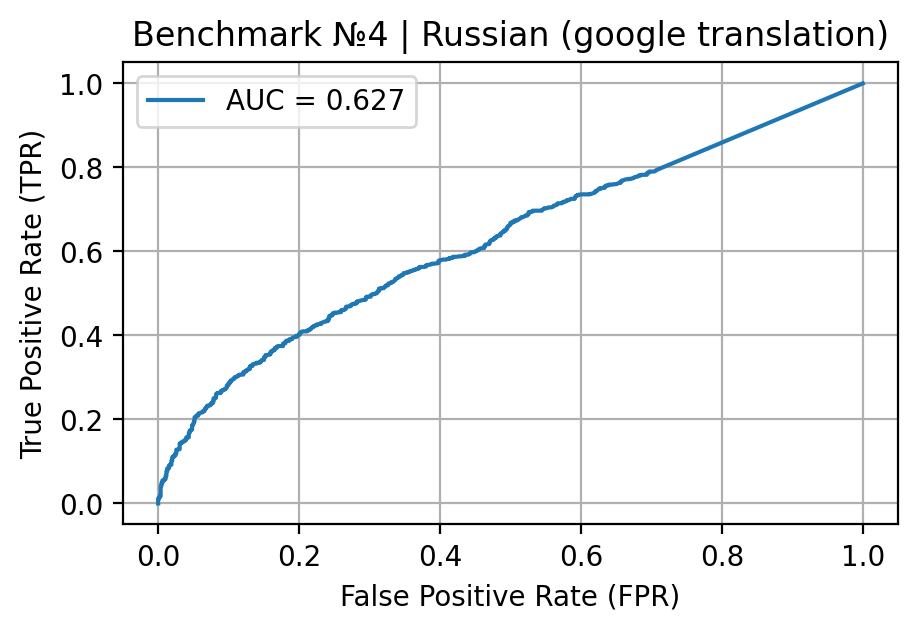}
\end{subfigure}
\caption{Representative ROC curves illustrating cross-lingual transfer (English vs.\ Russian; M2M100 vs.\ Google Translate) on Benchmark~1 (cleaner) and Benchmark~4 (noisier).}
\label{fig:roc_comparison}
\end{figure}

We quantify cross-lingual transfer using ROC-AUC across all benchmark--language pairs (Table~\ref{tab:auc_all}). Results reveal a clear benchmark-dependent pattern.
On curated prompt-injection benchmarks (Benchmarks~1--2), separability remains high across languages. In particular, Benchmark~2 achieves near-perfect AUC in English (0.993) and maintains strong performance under translation (0.847--0.884), indicating that similarity to an English codebook transfers reliably when attacks follow canonical jailbreak patterns.
In contrast, on behaviorally diverse benchmarks (Benchmarks~3--4), separability degrades substantially. AUC drops to 0.618--0.703 on Benchmark~3 and to 0.593--0.627 on Benchmark~4 across translations. These datasets contain more heterogeneous unsafe behaviors and less templated attack patterns, making similarity-based discrimination more challenging.
Figure~\ref{fig:roc_comparison} illustrates this contrast by comparing ROC curves from a cleaner setting (Benchmark~1) and a heterogeneous setting (Benchmark~4).

\begin{table}[hbt]
\centering
\caption{AUC-ROC scores across benchmarks and input languages (BGE-M3 embedder).}
\label{tab:auc_all}
\begin{tabular}{lccccccc}
\toprule
 & $\mathbf{eng}$ & $\mathbf{ru}_{\mathrm{m2m}}$ & $\mathbf{ru}_{\mathrm{gt}}$ & $\mathbf{zh}_{\mathrm{m2m}}$ & $\mathbf{zh}_{\mathrm{gt}}$ & $\mathbf{ar}_{\mathrm{m2m}}$ & $\mathbf{ar}_{\mathrm{gt}}$ \\
\midrule
$\mathbf{bench}_{1}$ & 0.829 & 0.785 & 0.806 & 0.781 & 0.792 & 0.765 & 0.782 \\
$\mathbf{bench}_{2}$ & 0.993 & 0.854 & 0.866 & 0.855 & 0.884 & 0.855 & 0.847 \\
$\mathbf{bench}_{3}$ & 0.618 & 0.675 & 0.694 & 0.694 & 0.703 & 0.660 & 0.703 \\
$\mathbf{bench}_{4}$ & 0.615 & 0.614 & 0.627 & 0.593 & 0.617 & 0.605 & 0.609 \\
\bottomrule
\end{tabular}
\end{table}

\paragraph{RQ2: Low-FPR regime and thresholding.}

Deployment-grade safety systems must operate under strict false-positive constraints. In Table~\ref{tab:bench1_fpr_leq_1}, we report the security-critical metric TPR at fixed FPR$\leq 1\%$, computed by identifying the per-language operating point that satisfies the constraint on the validation partition. To evaluate deployability without language-specific routing, we additionally fix a single global threshold $\tau_{\text{final}} = 0.66$, calibrated on the English validation split to approximate the FPR$\leq 1\%$ regime across translated queries. This global configuration enables zero-configuration inference while preserving strict false-positive bounds in the majority of cross-lingual settings.

Under this constraint, the contrast between the two regimes becomes even more pronounced. On Benchmark~2, similarity-based detection remains effective, achieving TPR between 78.5\% and 91.9\% across languages with BGE-M3. In this setting, canonical jailbreak templates remain sufficiently close to the English codebook even after translation.

On Benchmark~4, however, recall collapses to single digits (3.3--6.4\%) across all languages and embedding models. When FPR is tightly constrained, similarity-only filtering fails to capture the majority of unsafe prompts in this heterogeneous setting.

\begin{table}[hbt]
\centering
\caption{\textbf{Security-critical metric (FPR $\leq 1\%$):} TPR at fixed FPR$\leq 1\%$ by embedder, language/translation, and benchmark.}
\label{tab:bench1_fpr_leq_1}
\scalebox{0.85}{\begin{tabular}{lccccccc}
\toprule
\textbf{Embedder} & \textbf{Eng} & \textbf{Ru (m2m)} & \textbf{Ru (gt)} & \textbf{Zh (m2m)} & \textbf{Zh (gt)} & \textbf{Ar (m2m)} & \textbf{Ar (gt)} \\
\midrule
\multicolumn{8}{c}{\textbf{Benchmark 1}} \\
\midrule
bge-m3 & 25.6\% & 22.2\% & 22.2\% & 21.6\% & 22.8\% & 20.1\% & 21.1\% \\
multilingual-e5-large & 24.4\% & 14.7\% & 17.0\% & 4.8\% & 7.9\% & 14.9\% & 15.9\% \\
jina-embeddings-v3 & 12.4\% & 9.0\% & 8.8\% & 10.1\% & 9.9\% & 7.0\% & 7.0\% \\
\midrule
\multicolumn{8}{c}{\textbf{Benchmark 2}} \\
\midrule
bge-m3 & 91.9\% & 82.9\% & 85.3\% & 80.1\% & 87.7\% & 78.5\% & 82.3\% \\
multilingual-e5-large & 95.2\% & 82.3\% & 82.8\% & 74.8\% & 76.8\% & 81.6\% & 82.3\% \\
jina-embeddings-v3 & 93.8\% & 85.5\% & 88.6\% & 85.5\% & 91.3\% & 79.3\% & 85.6\% \\
\midrule
\multicolumn{8}{c}{\textbf{Benchmark 3}} \\
\midrule
bge-m3 & 23.0\% & 14.0\% & 18.0\% & 17.0\% & 21.0\% & 17.0\% & 21.0\% \\
multilingual-e5-large & 15.0\% & 10.0\% & 6.0\% & 9.0\% & 7.0\% & 10.0\% & 5.0\% \\
jina-embeddings-v3 & 19.0\% & 6.0\% & 6.0\% & 10.0\% & 13.0\% & 11.0\% & 16.0\% \\
\midrule
\multicolumn{8}{c}{\textbf{Benchmark 4}} \\
\midrule
bge-m3 & 3.3\% & 5.2\% & 5.8\% & 4.5\% & 5.3\% & 6.1\% & 6.4\% \\
multilingual-e5-large & 3.9\% & 6.1\% & 5.0\% & 1.7\% & 2.0\% & 6.4\% & 2.9\% \\
jina-embeddings-v3 & 3.8\% & 3.9\% & 6.5\% & 5.7\% & 6.0\% & 6.6\% & 7.3\% \\
\bottomrule
\end{tabular}}
\end{table}

\paragraph{RQ3: End-to-end jailbreak mitigation on target LLMs.}

We evaluate practical security impact by comparing successful jailbreaks without filtering (Attacks$_0$) and with the semantic codebook pre-filter (Attacks$_1$) across three target LLMs. Full per-language results are provided in Appendix~\ref{app:asr_results}; here we report aggregated statistics.

\begin{table}[h]
\centering
\caption{\textbf{Average relative reduction in successful jailbreaks (\%) across models and languages.}
Mean and standard deviation are computed over all model–language pairs within each benchmark.}
\label{tab:asr_reduction_summary}

\begin{tabular}{lcc}
\toprule
\textbf{Benchmark} & \textbf{Mean Reduction (\%)} & \textbf{Std} \\
\midrule
Benchmark 1 & 96.2 & $\pm$ 2.6 \\
Benchmark 2 & 50.0 & $\pm$ 17.4 \\
Benchmark 3 & 43.7 & $\pm$ 21.6 \\
Benchmark 4 & 18.6 & $\pm$ 13.8 \\
\bottomrule
\end{tabular}
\end{table}

\begin{table*}[t]
\centering
\small
\caption{Codebook subsampling on \textbf{Benchmark 1} (unsafe class). Larger codebooks improve TPR but increase FPR, highlighting a coverage--false-alarm trade-off.}
\label{tab:codebook_ablation_b1}
\begin{tabular}{l l c c c c}
\toprule
\textbf{Codebook Size} & \textbf{Language} & \textbf{TPR} & \textbf{FPR} & \textbf{TNR} & \textbf{FNR} \\
\midrule
25\%  & English (native)   & 0.9476 & 0.6316 & 0.3684 & 0.0524 \\
      & Russian (m2m)      & 0.6452 & 0.2318 & 0.7682 & 0.3548 \\
      & Chinese (m2m)      & 0.6195 & 0.2166 & 0.7834 & 0.3805 \\
      & Arabic (m2m)       & 0.5200 & 0.1558 & 0.8442 & 0.4800 \\
      & Russian (gt)       & 0.7713 & 0.3284 & 0.6716 & 0.2287 \\
      & Chinese (gt)       & 0.7438 & 0.3196 & 0.6804 & 0.2562 \\
      & Arabic (gt)        & 0.6618 & 0.2731 & 0.7269 & 0.3382 \\
\midrule
50\%  & English (native)   & 0.9638 & 0.7100 & 0.2900 & 0.0362 \\
      & Russian (m2m)      & 0.7200 & 0.3045 & 0.6955 & 0.2800 \\
      & Chinese (m2m)      & 0.6886 & 0.2753 & 0.7247 & 0.3114 \\
      & Arabic (m2m)       & 0.5952 & 0.2034 & 0.7966 & 0.4048 \\
      & Russian (gt)       & 0.8336 & 0.4196 & 0.5804 & 0.1664 \\
      & Chinese (gt)       & 0.8101 & 0.4009 & 0.5991 & 0.1899 \\
      & Arabic (gt)        & 0.7432 & 0.3383 & 0.6617 & 0.2568 \\
\midrule
75\%  & English (native)   & 0.9713 & 0.7368 & 0.2632 & 0.0287 \\
      & Russian (m2m)      & 0.7568 & 0.3506 & 0.6494 & 0.2432 \\
      & Chinese (m2m)      & 0.7182 & 0.3071 & 0.6929 & 0.2818 \\
      & Arabic (m2m)       & 0.6311 & 0.2313 & 0.7687 & 0.3689 \\
      & Russian (gt)       & 0.8651 & 0.4618 & 0.5382 & 0.1349 \\
      & Chinese (gt)       & 0.8399 & 0.4435 & 0.5565 & 0.1601 \\
      & Arabic (gt)        & 0.7717 & 0.3787 & 0.6213 & 0.2283 \\
\midrule
100\% & English (native)   & 0.9814 & 0.8012 & 0.1988 & 0.0186 \\
      & Russian (m2m)      & 0.8162 & 0.4385 & 0.5615 & 0.1838 \\
      & Chinese (m2m)      & 0.7798 & 0.3790 & 0.6210 & 0.2202 \\
      & Arabic (m2m)       & 0.6987 & 0.2904 & 0.7096 & 0.3013 \\
      & Russian (gt)       & 0.9077 & 0.5547 & 0.4453 & 0.0923 \\
      & Chinese (gt)       & 0.8862 & 0.5244 & 0.4756 & 0.1138 \\
      & Arabic (gt)        & 0.8346 & 0.4583 & 0.5417 & 0.1654 \\
\bottomrule
\end{tabular}
\end{table*}

\begin{table}[hbt]
\centering
\caption{\textbf{Overall discriminatory power (AUC-ROC).} AUC by embedder and language on each benchmark.}
\label{tab:bench1_auc_embedders}
\scalebox{0.85}{\begin{tabular}{lccccccc}
\toprule
\textbf{Embedder} & \textbf{Eng} & \textbf{Ru (m2m)} & \textbf{Ru (gt)} & \textbf{Zh (m2m)} & \textbf{Zh (gt)} & \textbf{Ar (m2m)} & \textbf{Ar (gt)} \\
\midrule
\multicolumn{8}{c}{\textbf{Benchmark 1}} \\
\midrule
bge-m3 & 0.830 & 0.785 & 0.810 & 0.781 & 0.792 & 0.765 & 0.782 \\
multilingual-e5-large & 0.852 & 0.763 & 0.765 & 0.733 & 0.730 & 0.712 & 0.720 \\
jina-embeddings-v3 & 0.767 & 0.707 & 0.718 & 0.726 & 0.727 & 0.709 & 0.698 \\
\midrule
\multicolumn{8}{c}{\textbf{Benchmark 2}} \\
\midrule
bge-m3 & 0.993 & 0.985 & 0.988 & 0.981 & 0.991 & 0.977 & 0.984 \\
multilingual-e5-large & 0.998 & 0.973 & 0.975 & 0.970 & 0.980 & 0.982 & 0.992 \\
jina-embeddings-v3 & 0.995 & 0.9864 & 0.989 & 0.987 & 0.991 & 0.972 & 0.9833 \\
\midrule
\multicolumn{8}{c}{\textbf{Benchmark 3}} \\
\midrule
bge-m3 & 0.719 & 0.673 & 0.693 & 0.693 & 0.702 & 0.661 & 0.704 \\
multilingual-e5-large & 0.708 & 0.685 & 0.713 & 0.687 & 0.669 & 0.705 & 0.734 \\
jina-embeddings-v3 & 0.700 & 0.717 & 0.737 & 0.713 & 0.733 & 0.698 & 0.749 \\
\midrule
\multicolumn{8}{c}{\textbf{Benchmark 4}} \\
\midrule
bge-m3 & 0.620 & 0.618 & 0.631 & 0.595 & 0.619 & 0.610 & 0.612 \\
multilingual-e5-large & 0.635 & 0.643 & 0.627 & 0.535 & 0.510 & 0.618 & 0.587 \\
jina-embeddings-v3 & 0.591 & 0.571 & 0.606 & 0.565 & 0.599 & 0.574 & 0.587 \\
\bottomrule
\end{tabular}}
\end{table}

Table~\ref{tab:asr_reduction_summary} summarizes the average relative reduction in successful jailbreaks across models and languages within each benchmark.
On canonical prompt-injection benchmarks (Benchmarks~1--2), the semantic filter removes a substantial fraction of successful attacks, achieving a mean reduction of 96.2\% on Benchmark~1 and 50.0\% on Benchmark~2. 
Under distribution shift, mitigation weakens significantly: the mean reduction drops to 43.7\% on Benchmark~3 and further to 18.6\% on the most heterogeneous benchmark (Benchmark~4). 

The increasing standard deviation across Benchmarks~2--4 indicates growing instability across models and languages in harder settings. These results closely mirror the separability and low-FPR findings from RQ1--RQ2: end-to-end mitigation is strong in canonical regimes but degrades sharply under heterogeneous unsafe distributions.
\textit{Note:} ASR$_1$ is computed over the subset of unsafe prompts that pass the filter. Consequently, ASR$_1$ may increase even when the absolute number of successful jailbreaks decreases. Absolute reduction therefore provides the more informative security metric.

\paragraph{Supporting analysis: codebook size and embedder choice.}

We additionally analyze (i) the sensitivity to codebook size and (ii) the sensitivity to the embedding model. Table~\ref{tab:codebook_ablation_b1} shows that larger codebooks tend to increase TPR but also substantially increase FPR, indicating a practical trade-off between coverage and false alarms. Tables~\ref{tab:bench1_auc_embedders} and~\ref{tab:bench1_fpr_leq_1} show that while AUC can be similar across embedders on some benchmarks, differences become pronounced in the low-FPR regime; for example, on Benchmark~1 (Chinese, m2m), BGE-M3 achieves TPR 21.6\% at FPR$\leq 1\%$ versus 4.8\% for multilingual-e5-large. On Benchmark~4, all embedders collapse to single-digit TPR at FPR$\leq 1\%$, reinforcing the limitation of similarity-only filtering in the hardest setting.

\section{Conclusion}

We examined whether cross-lingual jailbreak attacks can be mitigated using a fixed English semantic codebook combined with multilingual sentence embeddings. Our experiments reveal a clear two-regime pattern. 

On curated benchmarks dominated by canonical jailbreak templates, similarity-based detection transfers reliably across languages, achieving near-perfect separability (AUC up to 0.993) and high recall under strict false-positive constraints (TPR 78--92\% at FPR $\leq 1\%$). When deployed as an external pre-filter, it consistently reduces the absolute number of successful jailbreaks across target LLMs and translation pipelines. 

Under distribution shift, however, performance degrades substantially. On behaviorally diverse benchmarks, AUC drops to 0.59--0.63 and recall in the low-FPR regime falls to single digits. These results indicate that semantic similarity to a fixed English codebook effectively captures canonical attack patterns but does not generalize reliably to heterogeneous or adversarially rephrased unsafe content.

Our analysis further demonstrates that embedding choice and codebook size introduce critical trade-offs between coverage and false-alarm rates, particularly in the security-critical operating regime (Appendix~\ref{app:embedders_results}).While the proposed framework relies on a static English codebook—a design choice that ensures deployment stability but inherently limits adaptability to novel attack morphologies or language-specific harm formulations—it establishes a scalable, training-free first-line defense for black-box multilingual LLMs. Similarity-based filtering alone is insufficient for comprehensive protection and should be integrated into a broader, multi-layered safety pipeline.

\section{Future Work}

Our findings identify several specific  directions for extending semantic guardrails in multilingual settings.

\textbf{Codebook maintenance and adaptation.} To mitigate the static nature of the English codebook, future systems should implement a continuous update pipeline: (i) production false negatives are clustered in embedding space to identify emerging semantic centroids; (ii) candidate entries are validated via ensemble LLM adjudication and cross-lingual consistency checks; (iii) periodic codebook refreshes preserve coverage while controlling concept drift. This active-learning loop enables adaptation to evolving jailbreak tactics without full detector retraining.

\textbf{Native multilingual codebooks.} Extending the framework to incorporate language-specific templates—via controlled translation, semantic deduplication, and isotropy-aware alignment—may improve recall for typologically distant languages (e.g., Arabic, Chinese) and reduce reliance on translation-induced semantic drift.

\textbf{Hybrid and cascaded architectures.} Given the performance degradation under distribution shift, similarity-based filtering should be combined with orthogonal signals such as perplexity-based anomaly detection, syntactic pattern analysis, or LLM-verified secondary review. Cascaded pipelines can compensate for similarity degradation while maintaining strict operational FPR constraints.

\textbf{Input sensitivity and prompt structure.} Systematic analysis of how detection performance varies with input length, contextual complexity, and syntactic obfuscation will clarify the operational boundaries of embedding-based filtering and guide prompt-aware normalization strategies.

\textbf{Longitudinal deployment evaluation.} Real-world assessment on live traffic is necessary to quantify resilience to evolving attack distributions, language drift, and production constraints (latency, throughput, cost), ultimately establishing practical deployment guidelines for semantic guardrails.




\clearpage

\bibliography{references}


\clearpage
\appendix
\section{Dataset Statistics}
\label{app:datasets_info}

\begin{table}[htbp]
\caption{Datasets details by classes}
\label{tab:datasets-stats}
\small
\begin{tabular}{lrrrrl}
\toprule
\textbf{Dataset} & \textbf{Total} & \textbf{Safe} & \textbf{Unsafe} & \textbf{Lang.} \\
\midrule
jayavibhav/prompt-injection-safety & 60,000 & 27,971 & 32,029 & EN \\
xTRam1/safe-guard-prompt-injection & 10,300 & 7,150 & 3,150 & EN \\
JailbreakBench/JBB-Behaviors & 200 & 100 & 100 & EN \\
nvidia/Aegis-AI-Content-Safety-Dataset-2.0 & 33,416 & 13,773 & 19643 & EN \\
\bottomrule
\end{tabular}
\end{table}

\section{Metrics dependence by threshold} \label{app:tau_tables}
\begin{table}[h]
\centering
\caption{Threshold selection and performance metrics on Benchmark №1 across input languages and translation methods for three optimisation criteria. 
}
\label{tab:bench1_min_fpr}
\scalebox{0.9}{\begin{tabular}{lcccccc}

\toprule
{\textbf{Input Language}} & {\textbf{Threshold}} & {\textbf{TPR (\%)}} & {\textbf{FPR (\%)}} & {\textbf{TNR (\%)}} & {\textbf{FNR (\%)}} \\
\midrule
\multicolumn{6}{c}{\textbf{Thresholds for FPR $\leq$ 1\%}} \\
\midrule
\textbf{English (native)} & 0.790 & 27.2 & 1.3 & 98.6 & 72.8 \\[5pt]
\textbf{Russian (model)}  &  0.740 & 24.3 & 1.4 & 98.6 & 75.6 \\
\textbf{Russian (google)} &  0.750 & 24.2 & 1.3 & 98.6 & 75.7 \\[5pt]
\textbf{Chinese (model)}  &  0.740 & 23.0 & 1.5 & 98.4 & 76.9 \\
\textbf{Chinese (google)} & 0.750 & 24.7 & 1.5 & 98.4 & 75.3 \\[5pt]
\textbf{Arabic (model)}  & 0.730 & 23.5 & 1.7 & 98.2 & 76.5 \\
\textbf{Arabic (google)}  & 0.740 & 24.3 & 1.7 & 98.2 & 75.6 \\ [2pt]
\midrule
\multicolumn{6}{c}{\textbf{Best thresholds for maximising TPR \& TNR}} \\
\midrule
\textbf{English (native)} & 0.725 & 78.0 & 28.4 & 71.6 & 22.0 \\[5pt]
\textbf{Russian (model)}  &  0.670 & 75.5 & 34.5 & 65.5 & 24.5 \\
\textbf{Russian (google)} &  0.691 & 70.8 & 25.5 & 74.5 & 29.2 \\[5pt]
\textbf{Chinese (model)}  &  0.665 & 74.6 & 33.5 & 66.5 & 25.4 \\
\textbf{Chinese (google)} & 0.677 & 78.5 & 36.4 & 63.6 & 21.5 \\[5pt]
\textbf{Arabic (model)}  & 0.656 & 72.8 & 32.5 & 67.5 & 27.2 \\
\textbf{Arabic (google)}  & 0.670 & 76.6 & 36.5 & 63.5 & 23.4 \\ [2pt]
\midrule
\multicolumn{6}{c}{\textbf{Best thresholds for minimising FPR \& FNR}} \\
\midrule
\textbf{English (native)} & 0.729 & 75.3  & 28.4  & 71.6 & 22.0 \\[5pt]
\textbf{Russian (model)}  & 0.675 & 71.1 & 29.5 & 70.5 & 28.9 \\
\textbf{Russian (google)} & 0.691 & 70.8 & 25.5 & 74.5 & 29.2 \\[5pt]
\textbf{Chinese (model)}  & 0.670 & 70.8 & 29.4 & 70.6 & 29.2 \\
\textbf{Chinese (google)} & 0.686 & 72.3 & 29.4 & 70.6 & 27.7 \\[5pt]
\textbf{Arabic (model)}  & 0.659 & 71.0 & 30.4 & 69.6 & 29.0 \\
\textbf{Arabic (google)}  & 0.677 & 71.0 & 30.4 & 69.6 & 29.0 \\ [2pt]
\bottomrule
\end{tabular}
}
\end{table}

\begin{table}[htbp]
\centering
\caption{Threshold selection and performance metrics on Benchmark №2 across input languages and translation methods for three optimisation criteria. 
}
\label{tab:bench2_min_fpr}
\begin{tabular}{lcccccc}
\toprule
{\textbf{Input Language}} & {\textbf{Threshold}} & {\textbf{TPR (\%)}} & {\textbf{FPR (\%)}} & {\textbf{TNR (\%)}} & {\textbf{FNR (\%)}} \\
\midrule
\multicolumn{6}{c}{\textbf{Thresholds for FPR $\leq$ 1\%}} \\
\midrule
\textbf{English (native)}      & 0.600 & 98.7 & 0.0 & 99.6 & 1.3 \\[5pt]
\textbf{Russian (model)}       & 0.700 & 70.9 & 0.1 & 99.9 & 29.1 \\
\textbf{Russian (google)}      & 0.700 & 73.2 & 0.1 & 99.9 & 26.8 \\[5pt]
\textbf{Chinese (model)}       & 0.700 & 71.0 & 0.3 & 99.7 & 29.0 \\
\textbf{Chinese (google)}      & 0.700 & 76.8 & 0.2 & 99.8 & 23.2 \\[5pt]
\textbf{Arabic (model)}        & 0.700 & 71.0 & 0.3 & 99.7 & 29.0 \\
\textbf{Arabic (google)}       & 0.700 & 69.4 & 0.2 & 99.8 & 30.6 \\[2pt]
\midrule
\multicolumn{6}{c}{\textbf{Best thresholds for maximising TPR \& TNR}} \\
\midrule
\textbf{English (native)}      & 0.600 & 98.7 & 0.4 & 99.6 & 1.3 \\[5pt]
\textbf{Russian (model)}       & 0.700 & 70.9 & 0.1 & 99.9 & 29.1 \\
\textbf{Russian (google)}      & 0.700 & 73.2 & 0.1 & 99.9 & 26.8 \\[5pt]
\textbf{Chinese (model)}       & 0.700 & 71.0 & 0.3 & 99.7 & 29.0 \\
\textbf{Chinese (google)}      & 0.700 & 76.8 & 0.2 & 99.8 & 23.2 \\[5pt]
\textbf{Arabic (model)}        & 0.700 & 71.0 & 0.3 & 99.7 & 29.0 \\
\textbf{Arabic (google)}       & 0.700 & 69.4 & 0.2 & 99.8 & 30.6 \\[2pt]
\midrule
\multicolumn{6}{c}{\textbf{Best thresholds for minimising FPR \& FNR}} \\
\midrule
\textbf{English (native)}      & 0.600 & 98.7 & 0.0 & 99.6 & 1.3 \\[5pt]
\textbf{Russian (model)}       & 0.700 & 70.9 & 0.1 & 99.9 & 29.1 \\
\textbf{Russian (google)}      & 0.700 & 73.2 & 0.1 & 99.9 & 26.8 \\[5pt]
\textbf{Chinese (model)}       & 0.700 & 71.0 & 0.3 & 99.7 & 29.0 \\
\textbf{Chinese (google)}      & 0.700 & 76.8 & 0.2 & 99.8 & 23.2 \\[5pt]
\textbf{Arabic (model)}        & 0.700 & 71.0 & 0.3 & 99.7 & 29.0 \\
\textbf{Arabic (google)}       & 0.700 & 69.4 & 0.2 & 99.8 & 30.6 \\[2pt]
\bottomrule
\end{tabular}
\end{table}

\begin{table}[htbp]
\centering
\caption{Threshold selection and performance metrics on Benchmark №3 across input languages and translation methods for three optimisation criteria. 
}
\label{tab:bench3_min_fpr}
\begin{tabular}{lcccccc}
\toprule
{\textbf{Input Language}} & {\textbf{Threshold}} & {\textbf{TPR (\%)}} & {\textbf{FPR (\%)}} & {\textbf{TNR (\%)}} & {\textbf{FNR (\%)}} \\
\midrule
\multicolumn{6}{c}{\textbf{Thresholds for FPR $\leq$ 1\%}} \\
\midrule
\textbf{English (native)}      & 0.790 & 20.0 & 1.0 & 99.0 & 80.0 \\[5pt]
\textbf{Russian (model)}       & 0.780 & 12.0 & 1.0 & 99.0 & 88.0 \\
\textbf{Russian (google)}      & 0.750 & 18.0 & 1.0 & 99.0 & 82.0 \\[5pt]
\textbf{Chinese (model)}       & 0.770 & 15.0 & 1.0 & 99.0 & 85.0 \\
\textbf{Chinese (google)}      & 0.780 & 19.0 & 1.0 & 99.0 & 81.0 \\[5pt]
\textbf{Arabic (model)}        & 0.760 & 13.0 & 0.0 & 100.0 & 87.0 \\
\textbf{Arabic (google)}       & 0.750 & 16.0 & 0.0 & 100.0 & 84.0 \\[2pt]
\midrule
\multicolumn{6}{c}{\textbf{Best thresholds for maximising TPR \& TNR}} \\
\midrule
\textbf{English (native)}      & 0.714 & 41.0 & 17.0 & 83.0 & 59.0 \\[5pt]
\textbf{Russian (model)}       & 0.638 & 73.0 & 45.0 & 55.0 & 27.0 \\
\textbf{Russian (google)}      & 0.675 & 56.0 & 28.0 & 72.0 & 44.0 \\[5pt]
\textbf{Chinese (model)}       & 0.637 & 77.0 & 47.0 & 53.0 & 23.0 \\
\textbf{Chinese (google)}      & 0.665 & 67.0 & 37.0 & 63.0 & 33.0 \\[5pt]
\textbf{Arabic (model)}        & 0.647 & 64.0 & 39.0 & 61.0 & 36.0 \\
\textbf{Arabic (google)}       & 0.668 & 64.0 & 33.0 & 67.0 & 36.0 \\[2pt]
\midrule
\multicolumn{6}{c}{\textbf{Best thresholds for minimising FPR \& FNR}} \\
\midrule
\textbf{English (native)}      & 0.714 & 41.0 & 17.0 & 83.0 & 59.0 \\[5pt]
\textbf{Russian (model)}       & 0.652 & 64.0 & 38.0 & 62.0 & 36.0 \\
\textbf{Russian (google)}      & 0.675 & 56.0 & 28.0 & 72.0 & 44.0 \\[5pt]
\textbf{Chinese (model)}       & 0.661 & 64.0 & 36.0 & 64.0 & 36.0 \\
\textbf{Chinese (google)}      & 0.666 & 66.0 & 36.0 & 64.0 & 34.0 \\[5pt]
\textbf{Arabic (model)}        & 0.648 & 63.0 & 38.0 & 62.0 & 37.0 \\
\textbf{Arabic (google)}      & 0.668 & 64.0 & 33.0 & 67.0 & 36.0 \\[2pt]
\bottomrule
\end{tabular}
\end{table}

\begin{table}[htbp]
\centering
\caption{Threshold selection and performance metrics on Benchmark №4 across input languages and translation methods for three optimisation criteria. 
}
\label{tab:bench4_min_fpr}
\begin{tabular}{lcccccc}
\toprule
{\textbf{Input Language}} & {\textbf{Threshold}} & {\textbf{TPR (\%)}} & {\textbf{FPR (\%)}} & {\textbf{TNR (\%)}} & {\textbf{FNR (\%)}} \\
\midrule
\multicolumn{6}{c}{\textbf{Thresholds for FPR $\leq$ 1\%}} \\
\midrule
\textbf{English (native)}      & 0.780 & 5.1 & 1.0 & 98.6 & 94.9 \\[5pt]
\textbf{Russian (model)}       & 0.730 & 5.9 & 1.4 & 98.5 & 94.1 \\
\textbf{Russian (google)}      & 0.720 & 8.6 & 1.5 & 98.5 & 91.4 \\[5pt]
\textbf{Chinese (model)}       & 0.740 & 5.4 & 1.3 & 98.2 & 94.6 \\
\textbf{Chinese (google)}      & 0.730 & 8.4 & 1.5 & 98.5 & 91.6 \\[5pt]
\textbf{Arabic (model)}        & 0.720 & 7.1 & 1.3 & 98.7 & 92.9 \\
\textbf{Arabic (google)}       & 0.720 & 7.4 & 1.2 & 98.7 & 92.5 \\[2pt]
\midrule
\multicolumn{6}{c}{\textbf{Best thresholds for maximising TPR \& TNR}} \\
\midrule
\textbf{English (native)}      & 0.619 & 49.9 & 29.3 & 70.7 & 50.1 \\[5pt]
\textbf{Russian (model)}       & 0.622 & 36.6 & 17.5 & 82.5 & 63.4 \\
\textbf{Russian (google)}      & 0.615 & 44.7 & 24.4 & 75.6 & 55.3 \\[5pt]
\textbf{Chinese (model)}       & 0.603 & 48.5 & 33.5 & 66.5 & 51.5 \\
\textbf{Chinese (google)}      & 0.630 & 39.8 & 20.2 & 79.8 & 60.2 \\[5pt]
\textbf{Arabic (model)}        & 0.617 & 37.6 & 20.4 & 79.6 & 62.4 \\
\textbf{Arabic (google)}       & 0.620 & 39.6 & 20.4 & 79.6 & 60.4 \\[2pt]
\midrule
\multicolumn{6}{c}{\textbf{Best thresholds for minimising FPR \& FNR}} \\
\midrule
\textbf{English (native)}      & 0.619 & 49.9 & 29.3 & 70.7 & 50.1 \\[5pt]
\textbf{Russian (model)}       & 0.622 & 36.6 & 17.5 & 82.5 & 63.4 \\
\textbf{Russian (google)}      & 0.615 & 44.7 & 24.4 & 75.6 & 55.3 \\[5pt]
\textbf{Chinese (model)}       & 0.615 & 42.8 & 27.3 & 72.7 & 57.2 \\
\textbf{Chinese (google)}      & 0.629 & 40.8 & 21.5 & 78.5 & 59.2 \\[5pt]
\textbf{Arabic (model)}        & 0.617 & 37.6 & 20.4 & 79.6 & 62.4 \\
\textbf{Arabic (google)}       & 0.618 & 41.3 & 21.5 & 78.5 & 58.7 \\[2pt]
\bottomrule
\end{tabular}
\end{table}

\clearpage
\section{Full ASR Results}
\label{app:asr_results}

\begin{table}[htbp]
\centering
\caption{\textbf{Benchmark 1.} End-to-end mitigation results.
Attacks$_0$ denotes successful jailbreaks without filtering;
Attacks$_1$ denotes successful jailbreaks under defense.
$\Delta$Attacks = Attacks$_0$ - Attacks$_1$.}
\label{tab:bench1_asr_app}

\scalebox{0.88}{
\begin{tabular}{lccccc}
\toprule
\textbf{Input} & \textbf{Attacks$_0$} & \textbf{Attacks$_1$} & $\Delta$ \textbf{Attacks} & \textbf{ASR$_0$} & \textbf{ASR$_1$} \\
\midrule
\multicolumn{6}{c}{\textbf{Qwen3-4B}} \\
English (native) & 14 & 0 & 14 & 2.8\% & 0\% \\
Russian (model) & 214 & 5 & 209 & 42.8\% & 62.5\% \\
Russian (google) & 60 & 1 & 59 & 12.0\% & 25.0\% \\
Chinese (model) & 106 & 4 & 102 & 21.2\% & 66.7\% \\
Chinese (google) & 48 & 0 & 48 & 9.6\% & 0\% \\
Arabic (model) & 140 & 3 & 137 & 28.0\% & 21.4\% \\
Arabic (google) & 36 & 2 & 34 & 7.2\% & 20.0\% \\
\midrule
\multicolumn{6}{c}{\textbf{Llama-3.2-3B-Instruct}} \\
English (native) & 45 & 0 & 45 & 9.0\% & 0\% \\
Russian (model) & 131 & 6 & 125 & 26.2\% & 75.0\% \\
Russian (google) & 38 & 0 & 38 & 7.6\% & 0\% \\
Chinese (model) & 119 & 5 & 114 & 23.8\% & 83.3\% \\
Chinese (google) & 67 & 2 & 65 & 13.4\% & 40.0\% \\
Arabic (model) & 161 & 7 & 154 & 32.2\% & 50.0\% \\
Arabic (google) & 90 & 4 & 86 & 18.0\% & 40.0\% \\
\midrule
\multicolumn{6}{c}{\textbf{gpt-3.5-turbo}} \\
English (native) & 61 & 0 & 61 & 12.2\% & 0\% \\
Russian (model) & 92 & 5 & 87 & 18.4\% & 62.0\% \\
Russian (google) & 92 & 1 & 91 & 18.4\% & 25.0\% \\
Chinese (model) & 76 & 2 & 74 & 15.2\% & 33.3\% \\
Chinese (google) & 76 & 0 & 76 & 15.2\% & 0\% \\
Arabic (model) & 161 & 7 & 154 & 32.2\% & 50.0\% \\
Arabic (google) & 74 & 2 & 72 & 14.8\% & 20.0\% \\
\bottomrule
\end{tabular}}
\end{table}

\begin{table}[htbp]
\centering
\caption{\textbf{Benchmark 2.} End-to-end mitigation results. Attacks$_0$ denotes successful jailbreaks without filtering; Attacks$_1$ denotes successful jailbreaks under defense.
$\Delta$Attacks = Attacks$_0$ - Attacks$_1$.}
\label{tab:bench2_asr_app}

\scalebox{0.88}{
\begin{tabular}{lccccc}
\toprule
\textbf{Input} & \textbf{Attacks$_0$} & \textbf{Attacks$_1$} & $\Delta$ \textbf{Attacks} & \textbf{ASR$_0$} & \textbf{ASR$_1$} \\
\midrule
\multicolumn{6}{c}{\textbf{Qwen3-4B}} \\
English (native) & 38 & 5 & 33 & 7.6\% & 18.5\% \\
Russian (model) & 94 & 44 & 50 & 18.8\% & 31.0\% \\
Russian (google) & 49 & 23 & 26 & 9.8\% & 16.9\% \\
Chinese (model) & 73 & 29 & 44 & 14.6\% & 21.2\% \\
Chinese (google) & 41 & 23 & 18 & 8.2\% & 43.9\% \\
Arabic (model) & 70 & 30 & 40 & 14.0\% & 21.9\% \\
Arabic (google) & 37 & 20 & 17 & 7.4\% & 12.8\% \\
\midrule
\multicolumn{6}{c}{\textbf{Llama-3.2-3B-Instruct}} \\
English (native) & 75 & 3 & 72 & 15.0\% & 11.1\% \\
Russian (model) & 106 & 36 & 70 & 21.2\% & 23.4\% \\
Russian (google) & 69 & 28 & 41 & 13.8\% & 20.6\% \\
Chinese (model) & 114 & 41 & 73 & 22.8\% & 29.9\% \\
Chinese (google) & 77 & 25 & 52 & 15.4\% & 21.9\% \\
Arabic (model) & 136 & 49 & 87 & 27.2\% & 35.8\% \\
Arabic (google) & 53 & 12 & 41 & 10.6\% & 7.7\% \\
\midrule
\multicolumn{6}{c}{\textbf{gpt-3.5-turbo}} \\
English (native) & 45 & 9 & 36 & 9.0\% & 33.3\% \\
Russian (model) & 41 & 20 & 21 & 8.2\% & 14.1\% \\
Russian (google) & 41 & 19 & 22 & 8.2\% & 14.0\% \\
Chinese (model) & 31 & 13 & 18 & 6.2\% & 9.5\% \\
Chinese (google) & 31 & 9 & 22 & 6.2\% & 7.9\% \\
Arabic (model) & 136 & 49 & 87 & 27.2\% & 35.8\% \\
Arabic (google) & 39 & 19 & 20 & 7.8\% & 12.2\% \\
\bottomrule
\end{tabular}}
\end{table}

\begin{table}[htbp]
\centering
\caption{\textbf{Benchmark 3.} End-to-end mitigation results. Attacks$_0$ denotes successful jailbreaks without filtering; Attacks$_1$ denotes successful jailbreaks under defense.
$\Delta$Attacks = Attacks$_0$ - Attacks$_1$.}
\label{tab:bench3_asr_app}

\scalebox{0.88}{
\begin{tabular}{lccccc}
\toprule
\textbf{Input} & \textbf{Attacks$_0$} & \textbf{Attacks$_1$} & $\Delta$ \textbf{Attacks} & \textbf{ASR$_0$} & \textbf{ASR$_1$} \\
\midrule
\multicolumn{6}{c}{\textbf{Qwen3-4B}} \\
English & 6 & 5 & 1 & 1.2\% & 8.8\% \\
Russian (model) & 41 & 18 & 23 & 8.2\% & 42.9\% \\
Russian (google) & 24 & 14 & 10 & 4.8\% & 38.9\% \\
Chinese (model) & 28 & 13 & 15 & 5.6\% & 36.1\% \\
Chinese (google) & 16 & 7 & 9 & 3.2\% & 23.3\% \\
Arabic (model) & 37 & 24 & 13 & 7.4\% & 46.2\% \\
Arabic (google) & 17 & 6 & 11 & 3.4\% & 17.1\% \\
\midrule
\multicolumn{6}{c}{\textbf{Llama-3.2-3B-Instruct}} \\
English & 10 & 5 & 5 & 2.0\% & 8.8\% \\
Russian (model) & 32 & 11 & 21 & 6.4\% & 26.2\% \\
Russian (google) & 24 & 14 & 10 & 4.8\% & 38.9\% \\
Chinese (model) & 39 & 13 & 26 & 7.8\% & 36.1\% \\
Chinese (google) & 16 & 7 & 9 & 3.2\% & 23.3\% \\
Arabic (model) & 39 & 21 & 18 & 7.8\% & 40.4\% \\
Arabic (google) & 17 & 6 & 11 & 3.4\% & 17.1\% \\
\midrule
\multicolumn{6}{c}{\textbf{gpt-3.5-turbo}} \\
English & 22 & 14 & 8 & 4.4\% & 24.6\% \\
Russian (model) & 25 & 13 & 12 & 5.0\% & 31.0\% \\
Russian (google) & 25 & 12 & 13 & 5.0\% & 33.3\% \\
Chinese (model) & 34 & 15 & 19 & 6.8\% & 41.7\% \\
Chinese (google) & 34 & 13 & 21 & 6.8\% & 43.3\% \\
Arabic (model) & 39 & 21 & 18 & 7.8\% & 40.4\% \\
Arabic (google) & 21 & 10 & 11 & 4.2\% & 28.6\% \\
\bottomrule
\end{tabular}}
\end{table}

\begin{table}[htbp]
\centering
\caption{\textbf{Benchmark 4.} End-to-end mitigation results.
Attacks$_0$ denotes successful jailbreaks without filtering; Attacks$_1$ denotes successful jailbreaks under defense.
$\Delta$Attacks = Attacks$_0$ - Attacks$_1$.}
\label{tab:bench4_asr_app}

\scalebox{0.88}{
\begin{tabular}{lccccc}
\toprule
\textbf{Input} & \textbf{Attacks$_0$} & \textbf{Attacks$_1$} & $\Delta$ \textbf{Attacks} & \textbf{ASR$_0$} & \textbf{ASR$_1$} \\
\midrule
\multicolumn{6}{c}{\textbf{Qwen3-4B}} \\
English (native) & 46 & 35 & 11 & 9.2\% & 10.1\% \\
Russian (model) & 119 & 92 & 27 & 23.8\% & 23.4\% \\
Russian (google) & 66 & 51 & 15 & 13.2\% & 13.2\% \\
Chinese (model) & 107 & 88 & 19 & 21.4\% & 23.2\% \\
Chinese (google) & 44 & 36 & 8 & 8.8\% & 9.9\% \\
Arabic (model) & 69 & 60 & 9 & 13.8\% & 15.0\% \\
Arabic (google) & 74 & 61 & 13 & 14.8\% & 16.2\% \\
\midrule
\multicolumn{6}{c}{\textbf{Llama-3.2-3B-Instruct}} \\
English (native) & 46 & 35 & 11 & 9.2\% & 10.0\% \\
Russian (model) & 131 & 111 & 20 & 26.2\% & 28.2\% \\
Russian (google) & 66 & 51 & 15 & 13.2\% & 13.2\% \\
Chinese (model) & 107 & 88 & 19 & 21.4\% & 23.2\% \\
Chinese (google) & 44 & 36 & 8 & 8.8\% & 9.9\% \\
Arabic (model) & 115 & 60 & 55 & 23.0\% & 22.8\% \\
Arabic (google) & 74 & 61 & 13 & 14.8\% & 16.2\% \\
\midrule
\multicolumn{6}{c}{\textbf{gpt-3.5-turbo}} \\
English (native) & 39 & 26 & 13 & 7.8\% & 7.5\% \\
Russian (model) & 37 & 26 & 11 & 7.4\% & 6.6\% \\
Russian (google) & 37 & 27 & 10 & 7.4\% & 7.0\% \\
Chinese (model) & 34 & 22 & 12 & 6.8\% & 5.8\% \\
Chinese (google) & 34 & 22 & 12 & 6.8\% & 6.0\% \\
Arabic (model) & 115 & 91 & 24 & 23.0\% & 22.8\% \\
Arabic (google) & 35 & 23 & 12 & 7.0\% & 6.1\% \\
\bottomrule
\end{tabular}}
\end{table}

\clearpage
\section{Results of the Codebook Sizes}
\label{app:codebook_sized_results}

\begin{table}[ht]
\centering
\small
\caption{Detection performance across codebook subsampling ratios on \textbf{Benchmark 2}.}
\label{tab:codebook_ablation_b2}
\scalebox{0.9}{\begin{tabular}{l l c c c c}
\toprule
\textbf{Codebook Size} & \textbf{Language} & \textbf{TPR} & \textbf{FPR} & \textbf{TNR} & \textbf{FNR} \\
\midrule
25\%  & English (Original)   & 0.8242 & 0.0072 & 0.9928 & 0.1758 \\[2pt]
      & Russian (Model)      & 0.6107 & 0.0030 & 0.9970 & 0.3893 \\[2pt]
      & Chinese (Model)      & 0.6328 & 0.0051 & 0.9949 & 0.3672 \\[2pt]
      & Arabic (Model)       & 0.5668 & 0.0037 & 0.9963 & 0.4332 \\[2pt]
      & Russian (Google)     & 0.6377 & 0.0032 & 0.9968 & 0.3623 \\[2pt]
      & Chinese (Google)     & 0.7037 & 0.0042 & 0.9958 & 0.2963 \\[2pt]
      & Arabic (Google)      & 0.6193 & 0.0028 & 0.9972 & 0.3807 \\[2pt]
\midrule
50\%  & English (Original)   & 0.8906 & 0.0116 & 0.9884 & 0.1094 \\[2pt]
      & Russian (Model)      & 0.7176 & 0.0048 & 0.9952 & 0.2824 \\[2pt]
      & Chinese (Model)      & 0.7283 & 0.0077 & 0.9923 & 0.2717 \\[2pt]
      & Arabic (Model)       & 0.6865 & 0.0051 & 0.9949 & 0.3135 \\[2pt]
      & Russian (Google)     & 0.7525 & 0.0044 & 0.9956 & 0.2475 \\[2pt]
      & Chinese (Google)     & 0.7934 & 0.0062 & 0.9938 & 0.2066 \\[2pt]
      & Arabic (Google)      & 0.7205 & 0.0046 & 0.9954 & 0.2795 \\[2pt]
\midrule
75\%  & English (Original)   & 0.9082 & 0.0113 & 0.9887 & 0.0918 \\[2pt]
      & Russian (Model)      & 0.7451 & 0.0060 & 0.9940 & 0.2549 \\[2pt]
      & Chinese (Model)      & 0.7643 & 0.0083 & 0.9917 & 0.2357 \\[2pt]
      & Arabic (Model)       & 0.7102 & 0.0063 & 0.9937 & 0.2898 \\[2pt]
      & Russian (Google)     & 0.7746 & 0.0056 & 0.9944 & 0.2254 \\[2pt]
      & Chinese (Google)     & 0.8295 & 0.0062 & 0.9938 & 0.1705 \\[2pt]
      & Arabic (Google)      & 0.7508 & 0.0058 & 0.9942 & 0.2492 \\[2pt]
\midrule
100\% & English (Original)   & 0.9402 & 0.0183 & 0.9817 & 0.0598 \\[2pt]
      & Russian (Model)      & 0.7086 & 0.0012 & 0.9988 & 0.2914 \\[2pt]
      & Chinese (Model)      & 0.7102 & 0.0026 & 0.9974 & 0.2898 \\[2pt]
      & Arabic (Model)       & 0.7102 & 0.0026 & 0.9974 & 0.2898 \\[2pt]
      & Russian (Google)     & 0.8402 & 0.0088 & 0.9912 & 0.1598 \\[2pt]
      & Chinese (Google)     & 0.7684 & 0.0025 & 0.9975 & 0.2316 \\[2pt]
      & Arabic (Google)      & 0.6943 & 0.0016 & 0.9984 & 0.3057 \\[2pt]
\bottomrule
\end{tabular}
}
\end{table}

\begin{table}[htbp]
\centering
\small
\caption{Detection performance across codebook subsampling ratios on \textbf{Benchmark 3}.}
\label{tab:codebook_ablation_b3}
\begin{tabular}{l l c c c c}
\toprule
\textbf{Codebook Size} & \textbf{Language} & \textbf{TPR} & \textbf{FPR} & \textbf{TNR} & \textbf{FNR} \\
\midrule
25\%  & English (Original)   & 0.6200 & 0.2800 & 0.7200 & 0.3800 \\[2pt]
      & Russian (Model)      & 0.4300 & 0.1800 & 0.8200 & 0.5700 \\[2pt]
      & Chinese (Model)      & 0.4900 & 0.1700 & 0.8300 & 0.5100 \\[2pt]
      & Arabic (Model)       & 0.3400 & 0.1200 & 0.8800 & 0.6600 \\[2pt]
      & Russian (Google)     & 0.4700 & 0.1300 & 0.8700 & 0.5300 \\[2pt]
      & Chinese (Google)     & 0.4900 & 0.2000 & 0.8000 & 0.5100 \\[2pt]
      & Arabic (Google)      & 0.4800 & 0.1300 & 0.8700 & 0.5200 \\[2pt]
\midrule
50\%  & English (Original)   & 0.6800 & 0.4100 & 0.5900 & 0.3200 \\[2pt]
      & Russian (Model)      & 0.4700 & 0.1900 & 0.8100 & 0.5300 \\[2pt]
      & Chinese (Model)      & 0.4900 & 0.2700 & 0.7300 & 0.5100 \\[2pt]
      & Arabic (Model)       & 0.3700 & 0.2200 & 0.7800 & 0.6300 \\[2pt]
      & Russian (Google)     & 0.5200 & 0.2400 & 0.7600 & 0.4800 \\[2pt]
      & Chinese (Google)     & 0.6100 & 0.3300 & 0.6700 & 0.3900 \\[2pt]
      & Arabic (Google)      & 0.5100 & 0.2500 & 0.7500 & 0.4900 \\[2pt]
\midrule
75\%  & English (Original)   & 0.7300 & 0.4400 & 0.5600 & 0.2700 \\[2pt]
      & Russian (Model)      & 0.5100 & 0.2300 & 0.7700 & 0.4900 \\[2pt]
      & Chinese (Model)      & 0.5700 & 0.3000 & 0.7000 & 0.4300 \\[2pt]
      & Arabic (Model)       & 0.4100 & 0.2200 & 0.7800 & 0.5900 \\[2pt]
      & Russian (Google)     & 0.5800 & 0.2800 & 0.7200 & 0.4200 \\[2pt]
      & Chinese (Google)     & 0.6400 & 0.3700 & 0.6300 & 0.3600 \\[2pt]
      & Arabic (Google)      & 0.5500 & 0.2600 & 0.7400 & 0.4500 \\[2pt]
\midrule
100\% & English (Original)   & 0.8000 & 0.5500 & 0.4500 & 0.2000 \\[2pt]
      & Russian (Model)      & 0.5800 & 0.3600 & 0.6400 & 0.4200 \\[2pt]
      & Chinese (Model)      & 0.6400 & 0.3700 & 0.6300 & 0.3600 \\[2pt]
      & Arabic (Model)       & 0.4800 & 0.3200 & 0.6800 & 0.5200 \\[2pt]
      & Russian (Google)     & 0.6400 & 0.3900 & 0.6100 & 0.3600 \\[2pt]
      & Chinese (Google)     & 0.7000 & 0.4600 & 0.5400 & 0.3000 \\[2pt]
      & Arabic (Google)      & 0.6500 & 0.3700 & 0.6300 & 0.3500 \\[2pt]
\bottomrule
\end{tabular}
\end{table}

\begin{table}[htbp]
\centering
\small
\caption{Detection performance across codebook subsampling ratios on \textbf{Benchmark 4}.}
\label{tab:codebook_ablation_b4}
\begin{tabular}{l l c c c c}
\toprule
\textbf{Codebook Size} & \textbf{Language} & \textbf{TPR} & \textbf{FPR} & \textbf{TNR} & \textbf{FNR} \\
\midrule
25\%  & English (Original)   & 0.1686 & 0.0550 & 0.9450 & 0.8314 \\[2pt]
      & Russian (Model)      & 0.1156 & 0.0327 & 0.9673 & 0.8844 \\[2pt]
      & Chinese (Model)      & 0.1359 & 0.0329 & 0.9671 & 0.8641 \\[2pt]
      & Arabic (Model)       & 0.1108 & 0.0257 & 0.9743 & 0.8892 \\[2pt]
      & Russian (Google)     & 0.1350 & 0.0258 & 0.9742 & 0.8650 \\[2pt]
      & Chinese (Google)     & 0.1621 & 0.0352 & 0.9648 & 0.8379 \\[2pt]
      & Arabic (Google)      & 0.1417 & 0.0246 & 0.9754 & 0.8583 \\[2pt]
\midrule
50\%  & English (Original)   & 0.2158 & 0.0819 & 0.9181 & 0.7842 \\[2pt]
      & Russian (Model)      & 0.1484 & 0.0515 & 0.9485 & 0.8516 \\[2pt]
      & Chinese (Model)      & 0.1796 & 0.0599 & 0.9401 & 0.8204 \\[2pt]
      & Arabic (Model)       & 0.1435 & 0.0398 & 0.9602 & 0.8565 \\[2pt]
      & Russian (Google)     & 0.1641 & 0.0516 & 0.9484 & 0.8359 \\[2pt]
      & Chinese (Google)     & 0.1990 & 0.0704 & 0.9296 & 0.8010 \\[2pt]
      & Arabic (Google)      & 0.1699 & 0.0434 & 0.9566 & 0.8301 \\[2pt]
\midrule
75\%  & English (Original)   & 0.2331 & 0.0901 & 0.9099 & 0.7669 \\[2pt]
      & Russian (Model)      & 0.1638 & 0.0561 & 0.9439 & 0.8362 \\[2pt]
      & Chinese (Model)      & 0.1893 & 0.0599 & 0.9401 & 0.8107 \\[2pt]
      & Arabic (Model)       & 0.1522 & 0.0374 & 0.9626 & 0.8478 \\[2pt]
      & Russian (Google)     & 0.1816 & 0.0552 & 0.9448 & 0.8184 \\[2pt]
      & Chinese (Google)     & 0.2136 & 0.0681 & 0.9319 & 0.7864 \\[2pt]
      & Arabic (Google)      & 0.1845 & 0.0493 & 0.9507 & 0.8155 \\[2pt]
\midrule
100\% & English (Original)   & 0.2881 & 0.1181 & 0.8819 & 0.7119 \\[2pt]
      & Russian (Model)      & 0.2129 & 0.0749 & 0.9251 & 0.7871 \\[2pt]
      & Chinese (Model)      & 0.2350 & 0.0869 & 0.9131 & 0.7650 \\[2pt]
      & Arabic (Model)       & 0.2004 & 0.0667 & 0.9333 & 0.7996 \\[2pt]
      & Russian (Google)     & 0.2350 & 0.0739 & 0.9261 & 0.7650 \\[2pt]
      & Chinese (Google)     & 0.2612 & 0.0951 & 0.9049 & 0.7388 \\[2pt]
      & Arabic (Google)      & 0.2301 & 0.0751 & 0.9249 & 0.7699 \\[2pt]
\bottomrule
\end{tabular}
\end{table}

\clearpage
\section{Results of the Embedders Analysis}
\label{app:embedders_results}
\begin{table}[htb]
\centering
\footnotesize
\caption{Benchmark №2. Detailed threshold metrics for all models (separate translation methods). Part 1}
\label{tab:benchmark2-detailed_1}
\scalebox{0.8}{\begin{tabular}{llcccccccc}
\toprule
\textbf{Language} & \textbf{Model} & \textbf{AUC} & \textbf{Goal} & \textbf{Threshold} & \textbf{TPR} & \textbf{FPR} & \textbf{TNR} & \textbf{Precision} & \textbf{J/F1} \\
\midrule

\multirow{3}{*}{\textbf{English}} & \multirow{3}{*}{\textbf{bge-m3}} & \multirow{3}{*}{0.9929} & Youden's J & 0.6492 & 95.3\% & 2.6\% & 97.4\% & 94.1\% & J=0.927 \\
& & & Fixed FPR & 0.6789 & 91.9\% & 1.0\% & 99.0\% & 97.6\% & --- \\
& & & F1-Max & 0.6658 & 93.5\% & 1.4\% & 98.6\% & 96.6\% & F1=0.950 \\
\cmidrule{2-10}
& \multirow{3}{*}{\textbf{e5-large}} & \multirow{3}{*}{0.9976} & Youden's J & 0.8672 & 96.9\% & 1.9\% & 98.1\% & 95.7\% & J=0.950 \\
& & & Fixed FPR & 0.8736 & 95.2\% & 1.0\% & 99.0\% & 97.7\% & --- \\
& & & F1-Max & 0.8779 & 94.5\% & 0.4\% & 99.6\% & 99.0\% & F1=0.967 \\
\cmidrule{2-10}
& \multirow{3}{*}{\textbf{jina-v3}} & \multirow{3}{*}{0.9953} & Youden's J & 0.6916 & 97.0\% & 2.1\% & 97.9\% & 95.3\% & J=0.950 \\
& & & Fixed FPR & 0.7209 & 93.8\% & 1.0\% & 99.0\% & 97.6\% & --- \\
& & & F1-Max & 0.6966 & 96.4\% & 1.7\% & 98.3\% & 96.1\% & F1=0.963 \\
\midrule

\multirow{3}{*}{\textbf{Russian (model)}} & \multirow{3}{*}{\textbf{bge-m3}} & \multirow{3}{*}{0.9850} & Youden's J & 0.6139 & 93.1\% & 4.6\% & 95.4\% & 89.7\% & J=0.885 \\
& & & Fixed FPR & 0.6568 & 82.9\% & 1.0\% & 99.0\% & 97.4\% & --- \\
& & & F1-Max & 0.6274 & 90.7\% & 3.0\% & 97.0\% & 92.9\% & F1=0.918 \\
\cmidrule{2-10}
& \multirow{3}{*}{\textbf{e5-large}} & \multirow{3}{*}{0.9773} & Youden's J & 0.8425 & 89.3\% & 3.6\% & 96.4\% & 91.3\% & J=0.857 \\
& & & Fixed FPR & 0.8533 & 82.3\% & 1.0\% & 99.0\% & 97.3\% & --- \\
& & & F1-Max & 0.8449 & 87.9\% & 2.6\% & 97.4\% & 93.5\% & F1=0.906 \\
\cmidrule{2-10}
& \multirow{3}{*}{\textbf{jina-v3}} & \multirow{3}{*}{0.9864} & Youden's J & 0.6542 & 93.1\% & 3.4\% & 96.6\% & 92.1\% & J=0.896 \\
& & & Fixed FPR & 0.6871 & 85.5\% & 1.0\% & 99.0\% & 97.4\% & --- \\
& & & F1-Max & 0.6542 & 93.1\% & 3.4\% & 96.6\% & 92.1\% & F1=0.926 \\
\midrule

\multirow{3}{*}{\textbf{Russian (google)}} & \multirow{3}{*}{\textbf{bge-m3}} & \multirow{3}{*}{0.9876} & Youden's J & 0.6112 & 94.8\% & 5.2\% & 94.8\% & 88.6\% & J=0.896 \\
& & & Fixed FPR & 0.6558 & 85.3\% & 1.0\% & 99.0\% & 97.4\% & --- \\
& & & F1-Max & 0.6292 & 91.6\% & 2.7\% & 97.3\% & 93.5\% & F1=0.925 \\
\cmidrule{2-10}
& \multirow{3}{*}{\textbf{e5-large}} & \multirow{3}{*}{0.9749} & Youden's J & 0.8370 & 90.4\% & 5.2\% & 94.8\% & 88.2\% & J=0.852 \\
& & & Fixed FPR & 0.8508 & 82.8\% & 0.9\% & 99.1\% & 97.4\% & --- \\
& & & F1-Max & 0.8453 & 85.9\% & 1.9\% & 98.1\% & 95.1\% & F1=0.903 \\
\cmidrule{2-10}
& \multirow{3}{*}{\textbf{jina-v3}} & \multirow{3}{*}{0.9890} & Youden's J & 0.6424 & 95.7\% & 4.1\% & 95.9\% & 90.9\% & J=0.916 \\
& & & Fixed FPR & 0.6794 & 88.6\% & 1.0\% & 99.0\% & 97.5\% & --- \\
& & & F1-Max & 0.6573 & 92.8\% & 2.5\% & 97.5\% & 94.2\% & F1=0.935 \\
\midrule

\multirow{3}{*}{\textbf{Chinese (model)}} & \multirow{3}{*}{\textbf{bge-m3}} & \multirow{3}{*}{0.9810} & Youden's J & 0.6178 & 93.0\% & 5.7\% & 94.3\% & 87.5\% & J=0.873 \\
& & & Fixed FPR & 0.6674 & 80.1\% & 1.0\% & 99.0\% & 97.2\% & --- \\
& & & F1-Max & 0.6240 & 91.4\% & 4.5\% & 95.5\% & 89.7\% & F1=0.906 \\
\cmidrule{2-10}
& \multirow{3}{*}{\textbf{e5-large}} & \multirow{3}{*}{0.9696} & Youden's J & 0.8447 & 88.8\% & 8.0\% & 92.0\% & 82.7\% & J=0.808 \\
& & & Fixed FPR & 0.8543 & 74.8\% & 1.0\% & 99.0\% & 97.0\% & --- \\
& & & F1-Max & 0.8481 & 84.1\% & 4.3\% & 95.7\% & 89.3\% & F1=0.866 \\
\cmidrule{2-10}
& \multirow{3}{*}{\textbf{jina-v3}} & \multirow{3}{*}{0.9871} & Youden's J & 0.6678 & 94.1\% & 4.4\% & 95.6\% & 90.3\% & J=0.898 \\
& & & Fixed FPR & 0.7026 & 85.5\% & 1.0\% & 99.0\% & 97.4\% & --- \\
& & & F1-Max & 0.6747 & 93.0\% & 3.3\% & 96.7\% & 92.3\% & F1=0.927 \\
\bottomrule
\end{tabular}
}
\end{table}

\begin{table}[htb]
\centering
\footnotesize
\caption{Benchmark №2. Detailed threshold metrics for all models (separate translation methods). Part 2}
\label{tab:benchmark2-detailed_2}
\scalebox{0.8}{\begin{tabular}{llcccccccc}
\toprule
\textbf{Language} & \textbf{Model} & \textbf{AUC} & \textbf{Goal} & \textbf{Threshold} & \textbf{TPR} & \textbf{FPR} & \textbf{TNR} & \textbf{Precision} & \textbf{J/F1} \\
\midrule
\multirow{3}{*}{\textbf{Chinese (google)}} & \multirow{3}{*}{\textbf{bge-m3}} & \multirow{3}{*}{0.9906} & Youden's J & 0.6277 & 94.7\% & 3.6\% & 96.4\% & 91.9\% & J=0.911 \\
& & & Fixed FPR & 0.6617 & 87.7\% & 1.0\% & 99.0\% & 97.4\% & --- \\
& & & F1-Max & 0.6424 & 92.2\% & 2.0\% & 98.0\% & 95.1\% & F1=0.936 \\
\cmidrule{2-10}
& \multirow{3}{*}{\textbf{e5-large}} & \multirow{3}{*}{0.9796} & Youden's J & 0.8449 & 91.3\% & 7.1\% & 92.9\% & 84.7\% & J=0.843 \\
& & & Fixed FPR & 0.8545 & 76.8\% & 1.0\% & 99.0\% & 97.1\% & --- \\
& & & F1-Max & 0.8487 & 85.8\% & 3.7\% & 96.3\% & 90.9\% & F1=0.883 \\
\cmidrule{2-10}
& \multirow{3}{*}{\textbf{jina-v3}} & \multirow{3}{*}{0.9913} & Youden's J & 0.6785 & 95.2\% & 2.4\% & 97.6\% & 94.4\% & J=0.928 \\
& & & Fixed FPR & 0.6976 & 91.3\% & 1.0\% & 99.0\% & 97.5\% & --- \\
& & & F1-Max & 0.6818 & 95.0\% & 2.1\% & 97.9\% & 95.0\% & F1=0.950 \\
\midrule

\multirow{3}{*}{\textbf{Arabic (model)}} & \multirow{3}{*}{\textbf{bge-m3}} & \multirow{3}{*}{0.9773} & Youden's J & 0.6186 & 90.0\% & 5.3\% & 94.7\% & 87.9\% & J=0.847 \\
& & & Fixed FPR & 0.6577 & 78.5\% & 1.0\% & 99.0\% & 97.2\% & --- \\
& & & F1-Max & 0.6262 & 88.2\% & 3.9\% & 96.1\% & 90.8\% & F1=0.894 \\
\cmidrule{2-10}
& \multirow{3}{*}{\textbf{e5-large}} & \multirow{3}{*}{0.9822} & Youden's J & 0.8400 & 93.0\% & 5.1\% & 94.9\% & 88.7\% & J=0.879 \\
& & & Fixed FPR & 0.8546 & 81.6\% & 1.0\% & 99.0\% & 97.3\% & --- \\
& & & F1-Max & 0.8445 & 90.7\% & 3.3\% & 96.7\% & 92.1\% & F1=0.914 \\
\cmidrule{2-10}
& \multirow{3}{*}{\textbf{jina-v3}} & \multirow{3}{*}{0.9724} & Youden's J & 0.6343 & 90.5\% & 6.3\% & 93.7\% & 86.0\% & J=0.841 \\
& & & Fixed FPR & 0.6778 & 79.3\% & 1.0\% & 99.0\% & 97.2\% & --- \\
& & & F1-Max & 0.6549 & 85.7\% & 2.7\% & 97.3\% & 93.1\% & F1=0.893 \\
\midrule

\multirow{3}{*}{\textbf{Arabic (google)}} & \multirow{3}{*}{\textbf{bge-m3}} & \multirow{3}{*}{0.9844} & Youden's J & 0.6211 & 92.2\% & 3.9\% & 96.1\% & 91.0\% & J=0.883 \\
& & & Fixed FPR & 0.6578 & 82.3\% & 1.0\% & 99.0\% & 97.3\% & --- \\
& & & F1-Max & 0.6248 & 91.4\% & 3.4\% & 96.6\% & 92.1\% & F1=0.917 \\
\cmidrule{2-10}
& \multirow{3}{*}{\textbf{e5-large}} & \multirow{3}{*}{0.9915} & Youden's J & 0.8361 & 96.7\% & 5.5\% & 94.5\% & 88.4\% & J=0.913 \\
& & & Fixed FPR & 0.8558 & 84.8\% & 1.0\% & 99.0\% & 97.4\% & --- \\
& & & F1-Max & 0.8449 & 92.7\% & 2.5\% & 97.5\% & 94.1\% & F1=0.934 \\
\cmidrule{2-10}
& \multirow{3}{*}{\textbf{jina-v3}} & \multirow{3}{*}{0.9833} & Youden's J & 0.6402 & 92.1\% & 3.9\% & 96.1\% & 91.1\% & J=0.883 \\
& & & Fixed FPR & 0.6719 & 85.6\% & 1.0\% & 99.0\% & 97.4\% & --- \\
& & & F1-Max & 0.6524 & 90.2\% & 2.2\% & 97.8\% & 94.5\% & F1=0.923 \\
\bottomrule
\end{tabular}
}
\end{table}

\begin{table}[htbp]
\centering
\footnotesize
\caption{Benchmark №3. Detailed threshold metrics for all models (separate translation methods).}
\label{tab:benchmark3-detailed}
\scalebox{0.8}{\begin{tabular}{llcccccccc}
\toprule
\textbf{Language} & \textbf{Model} & \textbf{AUC} & \textbf{Goal} & \textbf{Threshold} & \textbf{TPR} & \textbf{FPR} & \textbf{TNR} & \textbf{Precision} & \textbf{J/F1} \\
\midrule

\multirow{3}{*}{\textbf{English}} & \multirow{3}{*}{\textbf{bge-m3}} & \multirow{3}{*}{0.7191} & Youden's J & 0.7148 & 56.0\% & 22.0\% & 78.0\% & 71.8\% & J=0.340 \\
& & & Fixed FPR & 0.8055 & 23.0\% & 1.0\% & 99.0\% & 95.8\% & --- \\
& & & F1-Max & 0.6181 & 93.0\% & 68.0\% & 32.0\% & 57.8\% & F1=0.713 \\
\cmidrule{2-10}
& \multirow{3}{*}{\textbf{e5-large}} & \multirow{3}{*}{0.7083} & Youden's J & 0.8980 & 59.0\% & 22.0\% & 78.0\% & 72.8\% & J=0.370 \\
& & & Fixed FPR & 0.9286 & 15.0\% & 0.0\% & 100.0\% & 100.0\% & --- \\
& & & F1-Max & 0.8826 & 80.0\% & 51.0\% & 49.0\% & 61.1\% & F1=0.693 \\
\cmidrule{2-10}
& \multirow{3}{*}{\textbf{jina-v3}} & \multirow{3}{*}{0.7000} & Youden's J & 0.7282 & 60.0\% & 28.0\% & 72.0\% & 68.2\% & J=0.320 \\
& & & Fixed FPR & 0.8414 & 19.0\% & 1.0\% & 99.0\% & 95.0\% & --- \\
& & & F1-Max & 0.6250 & 95.0\% & 78.0\% & 22.0\% & 54.9\% & F1=0.696 \\
\midrule

\multirow{3}{*}{\textbf{Russian (model)}} & \multirow{3}{*}{\textbf{bge-m3}} & \multirow{3}{*}{0.6732} & Youden's J & 0.6380 & 73.0\% & 45.0\% & 55.0\% & 61.9\% & J=0.280 \\
& & & Fixed FPR & 0.7759 & 14.0\% & 1.0\% & 99.0\% & 93.3\% & --- \\
& & & F1-Max & 0.6192 & 82.0\% & 57.0\% & 43.0\% & 59.0\% & F1=0.686 \\
\cmidrule{2-10}
& \multirow{3}{*}{\textbf{e5-large}} & \multirow{3}{*}{0.6846} & Youden's J & 0.8548 & 81.0\% & 52.0\% & 48.0\% & 60.9\% & J=0.290 \\
& & & Fixed FPR & 0.9006 & 10.0\% & 1.0\% & 99.0\% & 90.9\% & --- \\
& & & F1-Max & 0.8548 & 81.0\% & 52.0\% & 48.0\% & 60.9\% & F1=0.695 \\
\cmidrule{2-10}
& \multirow{3}{*}{\textbf{jina-v3}} & \multirow{3}{*}{0.7173} & Youden's J & 0.6691 & 53.0\% & 17.0\% & 83.0\% & 75.7\% & J=0.360 \\
& & & Fixed FPR & 0.7944 & 6.0\% & 0.0\% & 100.0\% & 100.0\% & --- \\
& & & F1-Max & 0.6037 & 84.0\% & 49.0\% & 51.0\% & 63.2\% & F1=0.721 \\
\midrule

\multirow{3}{*}{\textbf{Russian (google)}} & \multirow{3}{*}{\textbf{bge-m3}} & \multirow{3}{*}{0.6933} & Youden's J & 0.6764 & 54.0\% & 26.0\% & 74.0\% & 67.5\% & J=0.280 \\
& & & Fixed FPR & 0.7620 & 18.0\% & 0.0\% & 100.0\% & 100.0\% & --- \\
& & & F1-Max & 0.6085 & 87.0\% & 63.0\% & 37.0\% & 58.0\% & F1=0.696 \\
\cmidrule{2-10}
& \multirow{3}{*}{\textbf{e5-large}} & \multirow{3}{*}{0.7127} & Youden's J & 0.8706 & 58.0\% & 21.0\% & 79.0\% & 73.4\% & J=0.370 \\
& & & Fixed FPR & 0.9119 & 6.0\% & 0.0\% & 100.0\% & 100.0\% & --- \\
& & & F1-Max & 0.8519 & 89.0\% & 63.0\% & 37.0\% & 58.6\% & F1=0.706 \\
\cmidrule{2-10}
& \multirow{3}{*}{\textbf{jina-v3}} & \multirow{3}{*}{0.7369} & Youden's J & 0.6397 & 69.0\% & 25.0\% & 75.0\% & 73.4\% & J=0.440 \\
& & & Fixed FPR & 0.7896 & 6.0\% & 1.0\% & 99.0\% & 85.7\% & --- \\
& & & F1-Max & 0.6350 & 70.0\% & 26.0\% & 74.0\% & 72.9\% & F1=0.714 \\
\midrule

\multirow{3}{*}{\textbf{Chinese (model)}} & \multirow{3}{*}{\textbf{bge-m3}} & \multirow{3}{*}{0.6925} & Youden's J & 0.6374 & 77.0\% & 47.0\% & 53.0\% & 62.1\% & J=0.300 \\
& & & Fixed FPR & 0.7688 & 17.0\% & 1.0\% & 99.0\% & 94.4\% & --- \\
& & & F1-Max & 0.5880 & 93.0\% & 70.0\% & 30.0\% & 57.1\% & F1=0.707 \\
\cmidrule{2-10}
& \multirow{3}{*}{\textbf{e5-large}} & \multirow{3}{*}{0.6866} & Youden's J & 0.8388 & 77.0\% & 48.0\% & 52.0\% & 61.6\% & J=0.290 \\
& & & Fixed FPR & 0.8723 & 9.0\% & 1.0\% & 99.0\% & 90.0\% & --- \\
& & & F1-Max & 0.8369 & 82.0\% & 55.0\% & 45.0\% & 59.9\% & F1=0.692 \\
\cmidrule{2-10}
& \multirow{3}{*}{\textbf{jina-v3}} & \multirow{3}{*}{0.7129} & Youden's J & 0.6334 & 81.0\% & 47.0\% & 53.0\% & 63.3\% & J=0.340 \\
& & & Fixed FPR & 0.8114 & 10.0\% & 0.0\% & 100.0\% & 100.0\% & --- \\
& & & F1-Max & 0.6058 & 92.0\% & 64.0\% & 36.0\% & 59.0\% & F1=0.719 \\
\midrule

\multirow{3}{*}{\textbf{Chinese (google)}} & \multirow{3}{*}{\textbf{bge-m3}} & \multirow{3}{*}{0.7021} & Youden's J & 0.6661 & 66.0\% & 36.0\% & 64.0\% & 64.7\% & J=0.300 \\
& & & Fixed FPR & 0.7794 & 21.0\% & 1.0\% & 99.0\% & 95.5\% & --- \\
& & & F1-Max & 0.5818 & 97.0\% & 77.0\% & 23.0\% & 55.7\% & F1=0.708 \\
\cmidrule{2-10}
& \multirow{3}{*}{\textbf{e5-large}} & \multirow{3}{*}{0.6685} & Youden's J & 0.8477 & 55.0\% & 23.0\% & 77.0\% & 70.5\% & J=0.320 \\
& & & Fixed FPR & 0.8808 & 7.0\% & 0.0\% & 100.0\% & 100.0\% & --- \\
& & & F1-Max & 0.8203 & 99.0\% & 95.0\% & 5.0\% & 51.0\% & F1=0.674 \\
\cmidrule{2-10}
& \multirow{3}{*}{\textbf{jina-v3}} & \multirow{3}{*}{0.7333} & Youden's J & 0.6769 & 64.0\% & 29.0\% & 71.0\% & 68.8\% & J=0.350 \\
& & & Fixed FPR & 0.7906 & 13.0\% & 1.0\% & 99.0\% & 92.9\% & --- \\
& & & F1-Max & 0.6185 & 89.0\% & 57.0\% & 43.0\% & 61.0\% & F1=0.724 \\
\midrule

\multirow{3}{*}{\textbf{Arabic (model)}} & \multirow{3}{*}{\textbf{bge-m3}} & \multirow{3}{*}{0.6608} & Youden's J & 0.6480 & 63.0\% & 37.0\% & 63.0\% & 63.0\% & J=0.260 \\
& & & Fixed FPR & 0.7575 & 17.0\% & 0.0\% & 100.0\% & 100.0\% & --- \\
& & & F1-Max & 0.5472 & 98.0\% & 84.0\% & 16.0\% & 53.8\% & F1=0.695 \\
\cmidrule{2-10}
& \multirow{3}{*}{\textbf{e5-large}} & \multirow{3}{*}{0.7053} & Youden's J & 0.8622 & 61.0\% & 29.0\% & 71.0\% & 67.8\% & J=0.320 \\
& & & Fixed FPR & 0.8938 & 10.0\% & 1.0\% & 99.0\% & 90.9\% & --- \\
& & & F1-Max & 0.8509 & 85.0\% & 53.0\% & 47.0\% & 61.6\% & F1=0.714 \\
\cmidrule{2-10}
& \multirow{3}{*}{\textbf{jina-v3}} & \multirow{3}{*}{0.6981} & Youden's J & 0.6725 & 48.0\% & 16.0\% & 84.0\% & 75.0\% & J=0.320 \\
& & & Fixed FPR & 0.7761 & 11.0\% & 1.0\% & 99.0\% & 91.7\% & --- \\
& & & F1-Max & 0.5952 & 87.0\% & 63.0\% & 37.0\% & 58.0\% & F1=0.696 \\
\midrule

\multirow{3}{*}{\textbf{Arabic (google)}} & \multirow{3}{*}{\textbf{bge-m3}} & \multirow{3}{*}{0.7038} & Youden's J & 0.6682 & 64.0\% & 33.0\% & 67.0\% & 66.0\% & J=0.310 \\
& & & Fixed FPR & 0.7426 & 21.0\% & 1.0\% & 99.0\% & 95.5\% & --- \\
& & & F1-Max & 0.6248 & 84.0\% & 57.0\% & 43.0\% & 59.6\% & F1=0.697 \\
\cmidrule{2-10}
& \multirow{3}{*}{\textbf{e5-large}} & \multirow{3}{*}{0.7338} & Youden's J & 0.8620 & 71.0\% & 29.0\% & 71.0\% & 71.0\% & J=0.420 \\
& & & Fixed FPR & 0.9082 & 5.0\% & 0.0\% & 100.0\% & 100.0\% & --- \\
& & & F1-Max & 0.8582 & 79.0\% & 43.0\% & 57.0\% & 64.8\% & F1=0.712 \\
\cmidrule{2-10}
& \multirow{3}{*}{\textbf{jina-v3}} & \multirow{3}{*}{0.7494} & Youden's J & 0.6196 & 75.0\% & 31.0\% & 69.0\% & 70.8\% & J=0.440 \\
& & & Fixed FPR & 0.7415 & 16.0\% & 1.0\% & 99.0\% & 94.1\% & --- \\
& & & F1-Max & 0.5800 & 90.0\% & 56.0\% & 44.0\% & 61.6\% & F1=0.732 \\
\bottomrule
\end{tabular}
}
\end{table}

\begin{table}[htbp]
\centering
\footnotesize
\caption{Benchmark №4. Detailed threshold metrics for all models (separate translation methods).}
\label{tab:benchmark4-detailed}
\scalebox{0.8}{\begin{tabular}{llcccccccc}
\toprule
\textbf{Language} & \textbf{Model} & \textbf{AUC} & \textbf{Goal} & \textbf{Threshold} & \textbf{TPR} & \textbf{FPR} & \textbf{TNR} & \textbf{Precision} & \textbf{J/F1} \\
\midrule

\multirow{3}{*}{\textbf{English}} & \multirow{3}{*}{\textbf{bge-m3}} & \multirow{3}{*}{0.6199} & Youden's J & 0.6186 & 49.9\% & 29.2\% & 70.8\% & 67.4\% & J=0.207 \\
& & & Fixed FPR & 0.8050 & 3.3\% & 0.9\% & 99.1\% & 81.0\% & --- \\
& & & F1-Max & 0.4734 & 98.9\% & 95.8\% & 4.2\% & 55.6\% & F1=0.712 \\
\cmidrule{2-10}
& \multirow{3}{*}{\textbf{e5-large}} & \multirow{3}{*}{0.6345} & Youden's J & 0.8496 & 50.5\% & 29.5\% & 70.5\% & 67.5\% & J=0.210 \\
& & & Fixed FPR & 0.9031 & 3.9\% & 0.9\% & 99.1\% & 83.7\% & --- \\
& & & F1-Max & 0.7978 & 99.8\% & 98.2\% & 1.8\% & 55.2\% & F1=0.711 \\
\cmidrule{2-10}
& \multirow{3}{*}{\textbf{jina-v3}} & \multirow{3}{*}{0.5907} & Youden's J & 0.6785 & 31.8\% & 13.5\% & 86.5\% & 74.2\% & J=0.183 \\
& & & Fixed FPR & 0.8131 & 3.8\% & 0.9\% & 99.1\% & 83.0\% & --- \\
& & & F1-Max & 0.4842 & 99.1\% & 97.4\% & 2.6\% & 55.3\% & F1=0.710 \\
\midrule

\multirow{3}{*}{\textbf{Russian (model)}} & \multirow{3}{*}{\textbf{bge-m3}} & \multirow{3}{*}{0.6178} & Youden's J & 0.6222 & 36.6\% & 17.2\% & 82.8\% & 72.1\% & J=0.194 \\
& & & Fixed FPR & 0.7366 & 5.2\% & 0.9\% & 99.1\% & 87.1\% & --- \\
& & & F1-Max & 0.4627 & 98.7\% & 95.4\% & 4.6\% & 55.7\% & F1=0.712 \\
\cmidrule{2-10}
& \multirow{3}{*}{\textbf{e5-large}} & \multirow{3}{*}{0.6428} & Youden's J & 0.8322 & 50.7\% & 26.9\% & 73.1\% & 69.6\% & J=0.238 \\
& & & Fixed FPR & 0.8705 & 6.1\% & 0.9\% & 99.1\% & 88.7\% & --- \\
& & & F1-Max & 0.7532 & 100.0\% & 98.9\% & 1.1\% & 55.1\% & F1=0.711 \\
\cmidrule{2-10}
& \multirow{3}{*}{\textbf{jina-v3}} & \multirow{3}{*}{0.5708} & Youden's J & 0.6440 & 30.2\% & 16.8\% & 83.2\% & 68.5\% & J=0.133 \\
& & & Fixed FPR & 0.7612 & 3.9\% & 0.9\% & 99.1\% & 83.7\% & --- \\
& & & F1-Max & 0.4798 & 99.4\% & 98.4\% & 1.6\% & 55.1\% & F1=0.709 \\
\midrule

\multirow{3}{*}{\textbf{Russian (google)}} & \multirow{3}{*}{\textbf{bge-m3}} & \multirow{3}{*}{0.6305} & Youden's J & 0.6222 & 40.9\% & 20.2\% & 79.8\% & 71.0\% & J=0.207 \\
& & & Fixed FPR & 0.7435 & 5.8\% & 0.9\% & 99.1\% & 88.2\% & --- \\
& & & F1-Max & 0.4736 & 98.7\% & 95.8\% & 4.2\% & 55.5\% & F1=0.710 \\
\cmidrule{2-10}
& \multirow{3}{*}{\textbf{e5-large}} & \multirow{3}{*}{0.6265} & Youden's J & 0.8371 & 42.8\% & 22.7\% & 77.3\% & 69.6\% & J=0.202 \\
& & & Fixed FPR & 0.8737 & 5.0\% & 0.9\% & 99.1\% & 86.4\% & --- \\
& & & F1-Max & 0.7486 & 100.0\% & 99.1\% & 0.9\% & 55.0\% & F1=0.709 \\
\cmidrule{2-10}
& \multirow{3}{*}{\textbf{jina-v3}} & \multirow{3}{*}{0.6063} & Youden's J & 0.6551 & 29.5\% & 11.2\% & 88.8\% & 76.2\% & J=0.184 \\
& & & Fixed FPR & 0.7455 & 6.5\% & 0.9\% & 99.1\% & 89.3\% & --- \\
& & & F1-Max & 0.4621 & 100.0\% & 98.7\% & 1.3\% & 55.1\% & F1=0.710 \\
\midrule

\multirow{3}{*}{\textbf{Chinese (model)}} & \multirow{3}{*}{\textbf{bge-m3}} & \multirow{3}{*}{0.5952} & Youden's J & 0.6342 & 33.9\% & 17.8\% & 82.2\% & 69.7\% & J=0.160 \\
& & & Fixed FPR & 0.7554 & 4.5\% & 0.8\% & 99.2\% & 86.8\% & --- \\
& & & F1-Max & 0.4348 & 99.7\% & 98.6\% & 1.4\% & 55.0\% & F1=0.709 \\
\cmidrule{2-10}
& \multirow{3}{*}{\textbf{e5-large}} & \multirow{3}{*}{0.5346} & Youden's J & 0.8220 & 62.9\% & 54.9\% & 45.1\% & 58.1\% & J=0.080 \\
& & & Fixed FPR & 0.8675 & 1.7\% & 0.8\% & 99.2\% & 70.8\% & --- \\
& & & F1-Max & 0.7778 & 100.0\% & 99.2\% & 0.8\% & 54.9\% & F1=0.709 \\
\cmidrule{2-10}
& \multirow{3}{*}{\textbf{jina-v3}} & \multirow{3}{*}{0.5647} & Youden's J & 0.6771 & 21.6\% & 10.2\% & 89.8\% & 71.8\% & J=0.113 \\
& & & Fixed FPR & 0.7433 & 5.7\% & 0.9\% & 99.1\% & 88.1\% & --- \\
& & & F1-Max & 0.4902 & 99.8\% & 98.8\% & 1.2\% & 55.0\% & F1=0.709 \\
\midrule

\multirow{3}{*}{\textbf{Chinese (google)}} & \multirow{3}{*}{\textbf{bge-m3}} & \multirow{3}{*}{0.6194} & Youden's J & 0.6301 & 39.9\% & 20.2\% & 79.8\% & 70.5\% & J=0.197 \\
& & & Fixed FPR & 0.7556 & 5.3\% & 0.7\% & 99.3\% & 90.2\% & --- \\
& & & F1-Max & 0.4525 & 99.6\% & 98.5\% & 1.5\% & 55.0\% & F1=0.709 \\
\cmidrule{2-10}
& \multirow{3}{*}{\textbf{e5-large}} & \multirow{3}{*}{0.5059} & Youden's J & 0.8134 & 79.2\% & 74.6\% & 25.4\% & 56.2\% & J=0.046 \\
& & & Fixed FPR & 0.8751 & 2.0\% & 0.8\% & 99.2\% & 75.0\% & --- \\
& & & F1-Max & 0.7743 & 100.0\% & 99.5\% & 0.5\% & 54.8\% & F1=0.708 \\
\cmidrule{2-10}
& \multirow{3}{*}{\textbf{jina-v3}} & \multirow{3}{*}{0.5990} & Youden's J & 0.6260 & 52.0\% & 37.4\% & 62.6\% & 62.7\% & J=0.146 \\
& & & Fixed FPR & 0.7610 & 6.0\% & 0.8\% & 99.2\% & 89.9\% & --- \\
& & & F1-Max & 0.5039 & 98.9\% & 97.1\% & 2.9\% & 55.2\% & F1=0.709 \\
\midrule

\multirow{3}{*}{\textbf{Arabic (model)}} & \multirow{3}{*}{\textbf{bge-m3}} & \multirow{3}{*}{0.6102} & Youden's J & 0.6296 & 32.7\% & 15.0\% & 85.0\% & 72.6\% & J=0.177 \\
& & & Fixed FPR & 0.7272 & 6.1\% & 0.9\% & 99.1\% & 88.7\% & --- \\
& & & F1-Max & 0.4349 & 99.6\% & 97.9\% & 2.1\% & 55.3\% & F1=0.711 \\
\cmidrule{2-10}
& \multirow{3}{*}{\textbf{e5-large}} & \multirow{3}{*}{0.6180} & Youden's J & 0.8285 & 44.4\% & 26.2\% & 73.8\% & 67.3\% & J=0.182 \\
& & & Fixed FPR & 0.8659 & 6.4\% & 0.8\% & 99.2\% & 90.4\% & --- \\
& & & F1-Max & 0.7446 & 100.0\% & 98.4\% & 1.6\% & 55.2\% & F1=0.712 \\
\cmidrule{2-10}
& \multirow{3}{*}{\textbf{jina-v3}} & \multirow{3}{*}{0.5744} & Youden's J & 0.6472 & 25.4\% & 11.0\% & 89.0\% & 73.7\% & J=0.144 \\
& & & Fixed FPR & 0.7251 & 6.6\% & 0.9\% & 99.1\% & 89.5\% & --- \\
& & & F1-Max & 0.4377 & 100.0\% & 99.9\% & 0.1\% & 54.9\% & F1=0.709 \\
\midrule

\multirow{3}{*}{\textbf{Arabic (google)}} & \multirow{3}{*}{\textbf{bge-m3}} & \multirow{3}{*}{0.6121} & Youden's J & 0.6184 & 41.3\% & 21.4\% & 78.6\% & 70.0\% & J=0.199 \\
& & & Fixed FPR & 0.7273 & 6.4\% & 0.8\% & 99.2\% & 90.4\% & --- \\
& & & F1-Max & 0.4457 & 99.2\% & 97.4\% & 2.6\% & 55.2\% & F1=0.709 \\
\cmidrule{2-10}
& \multirow{3}{*}{\textbf{e5-large}} & \multirow{3}{*}{0.5871} & Youden's J & 0.8258 & 47.7\% & 33.5\% & 66.5\% & 63.3\% & J=0.142 \\
& & & Fixed FPR & 0.8783 & 2.9\% & 0.8\% & 99.2\% & 81.1\% & --- \\
& & & F1-Max & 0.7520 & 99.7\% & 98.4\% & 1.6\% & 55.1\% & F1=0.710 \\
\cmidrule{2-10}
& \multirow{3}{*}{\textbf{jina-v3}} & \multirow{3}{*}{0.5871} & Youden's J & 0.6414 & 31.8\% & 16.2\% & 83.8\% & 70.4\% & J=0.157 \\
& & & Fixed FPR & 0.7271 & 7.3\% & 0.9\% & 99.1\% & 90.4\% & --- \\
& & & F1-Max & 0.4456 & 100.0\% & 99.8\% & 0.2\% & 54.8\% & F1=0.708 \\
\bottomrule
\end{tabular}
}
\end{table}

\end{document}